\documentclass[journal]{IEEEtran}
\usepackage{blindtext}
\usepackage{amsmath}
\usepackage[export]{adjustbox}
\usepackage{graphicx}
\usepackage{diagbox}
\usepackage[table,xcdraw]{xcolor}
\usepackage{url}

\usepackage[]{microtype} 

\usepackage{cite} 

\usepackage[normalem]{ulem}
\useunder{\uline}{\ul}{}

\usepackage{framed,multirow}

\usepackage{comment}
\usepackage{lscape}
\usepackage[table,xcdraw]{xcolor}
\usepackage{makecell}
\usepackage{hyperref}

\ifCLASSINFOpdf
\else
\fi
\usepackage{caption}
\usepackage[caption=false]{subfig}
\begin{document}

%
\title{
CP\_HDR: A feature point detection and description library for LDR and HDR images
}

\author{
\IEEEauthorblockN{Artur Santos Nascimento*, Valter Guilherme Silva de Souza, \\ Daniel Oliveira Dantas, Beatriz Trinchão Andrade }\\
\IEEEauthorblockA{Departamento de Computação, Universidade Federal de Sergipe (UFS), São Cristóvão, Brazil\\
\{artur.nascimento, vgssouza, ddantas, beatriz\}@dcomp.ufs.br}
}
\maketitle


\begin{abstract}
In computer vision, \textit{characteristics} refer to image regions with unique properties, such as corners, edges, textures, or areas with high contrast. These regions can be represented through feature points (FPs). FP detection and description are fundamental steps to many computer vision tasks. Most FP detection and description methods use low dynamic range (LDR) images, sufficient for most applications involving digital images. However, LDR images may have saturated pixels in scenes with extreme light conditions, which degrade FP detection. On the other hand, high dynamic range (HDR) images usually present a greater dynamic range but FP detection algorithms do not take advantage of all the information in such images. In this study, we present a systematic review of image detection and description algorithms that use HDR images as input. We developed a library called CP\_HDR that implements the Harris corner detector, SIFT detector and descriptor, and two modifications of those algorithms specialized in HDR images, called SIFT for HDR (SfHDR) and Harris for HDR (HfHDR). Previous studies investigated the use of HDR images in FP detection, but we did not find studies investigating the use of HDR images in FP description. Using uniformity, repeatability rate, mean average precision, and matching rate metrics, we compared the performance of the CP\_HDR algorithms using LDR and HDR images. We observed an increase in the uniformity of the distribution of FPs among the high-light, mid-light, and low-light areas of the images. The results show that using HDR images as input to detection algorithms improves performance and that SfHDR and HfHDR enhance FP description.
\end{abstract}


\IEEEpeerreviewmaketitle


\section{Introduction}
     Several computer vision applications, such as 3D reconstruction, face recognition, image stitching, and object tracking, rely on feature points (FPs) detection and description~\cite{schmid2000evaluation, szeliski2010computerVision, chermak2014, andrade20123d}. Most FP detection algorithms are designed for low dynamic range (LDR) images. When a scene has extreme light conditions (with very dark and bright areas), the FP detector performance decreases due to under and overexposed areas in the image, missing potential FPs. High dynamic range (HDR) images can be used to overcome these problems~\cite{rana2015}.
    
     FP extraction algorithms detect and describe characteristics in images. The detection aims to find FPs in regions that represent image features. Good FP extraction algorithms detect FPs robust to image transformations such as rotation, point of view, and scale. The description involves assigning a description vector that identify the detected FP. An ideal description allows the identification of a certain FP even after several image transformations or different captures of the same scene~\cite{aguilera2012multispectral, hassaballah2016}.

     One of the major challenges in this area is to extract FPs from images with extreme lighting conditions, where low dynamic range (LDR) images can not accurately register and some areas of the image may end up being under or overexposed. The range of values that can be represented in an image is called dynamic range. While LDR image pixels usually can represent $2^8$ tones per sample~\cite{gonzalez}, HDR images are those whose pixels can represent more than $2^8$ tones (generally, $2^{32}$ tones) per sample. 
     
     
     If the difference between a scene's brightest and darkest regions is greater than its dynamic range, the brightest or darkest regions can be noisy, losing details. As HDR images have a greater dynamic range than LDR images, HDR can better represent scenes with extreme lighting conditions, especially in underexposed and overexposed areas~\cite{tm_reinhard2010}.

     Since FP extraction depends on the scene's lighting, using HDR images in FPs extraction can increase the number of FPs detected in saturated regions in LDR images~\cite{pvribyl2012, rana2015, pvribyl2016, welerson_hdr}. Furthermore, employing descriptors that enable a reliable description of these points can improve the performance of applications that use FP description.

     However, most FP extraction algorithms are designed to receive LDR images as input, and modifying the canonical algorithms to receive HDR images is a laborious task. Most studies that use HDR, use tone mapping (TM) algorithms to transform HDR into LDR images~\cite{pvribyl2012, pvribyl2016, rana2015, zhuang2019, mukherjee2021}. In previous studies, we have shown that using HDR images instead of LDR as input to FP extraction algorithms lead to better detection and description metrics~\cite{welerson_hdr, nascimento2022, nascimento2023}.

     In this study, we propose a library called CP\_HDR that can receive both LDR and HDR images as input to detection and description algorithms. We use the CP\_HDR library to compare the detection and description algorithms performance when using LDR and HDR images as input. The CP\_HDR is designed to be easy-to-use and popularize the FP extraction with HDR images as input, providing both canonical and HDR-optimized FP extraction algorithm. We made a systematic review to understand the state-of-the-art and list the datasets, algorithms, and metrics that are being used in the literature (Section~\ref{sec:systematic-review}); based on the systematic review results, we have implemented the CP\_HDR library. The metrics, datasets, algorithms and execution pipeline implemented are described in Section~\ref{sec:methods}. Afterwards, we compared our implementation with previous studies, showing the results (Section~\ref{sec:results}) and conclusions (Section~\ref{sec:conclusion}).


\section{Systematic review}~\label{sec:systematic-review}

This section describes a systematic review (SR) about FP detection and description algorithms using HDR images as input. The SR consists of three steps: \textit{planning}, \textit{conduction}, and \textit{description}. In the planning step, the main objective and protocol are defined. In the conduction step, the studies are searched and selected. In the description step, the relevant information is extracted and described~\cite{mapeamento, revisao}.

\subsection{Planning}~\label{subsec:sr-planning}
        
The main objective of this SR is to find relevant studies that explore FP detection and description algorithms that use HDR images as input or detail how detection and description in HDR images are made, comparing the algorithms performance when using HDR and LDR images. With that in mind, we formulated four search questions (SQs). \textbf{SQ1}: Which are the tone mapping techniques used? \textbf{SQ2}: Which are the detection and description algorithms used? \textbf{SQ3}: Which are the evaluation metrics used? \textbf{SQ4}: Which are the datasets used?
        
To find peer-reviewed studies, we used the following search sources: IEEE Xplore, ACM Digital Library, Elsevier Scopus, Springer Link, and Web of Science. The search string used was: \textit{( `Keypoint' OR `Feature Point' OR `Feature detection' ) AND ( `detection' OR `detector' OR `description' OR `descriptor' ) AND ( `High Dynamic Range' OR `HDR' ) AND `Image'}. The search string slightly changed depending on the search source due to syntax differences.

Three inclusion criteria (IC) and five exclusion criteria (EC) were used to define if a study is selected or discarded. \textbf{IC1}: the study proposes an approach for detecting or describing FPs in HDR images. \textbf{IC2}: the study describes the development of an FP detection or description algorithm that receives an HDR image as input. \textbf{IC3}: the study compares FP detectors or descriptors using LDR and HDR images as input. \textbf{EC1}: the study is not in Portuguese or English languages. \textbf{EC2}: only the abstract is available. \textbf{EC3}: the study is not accessible. \textbf{EC4}: no inclusion criteria were detected. And \textbf{EC5}: is not a primary source (reviews, mappings, surveys, tutorials etc).

\subsection{Conduction}~\label{subsec:sr-conduction}
        
The conduction step used the search string to find studies in the search sources. The duplicated and unavailable studies were discarded. Afterwards, the studies were read, and those who met any IC and no EC were selected to extract relevant information.

\begin{figure}[!htb]
            \centering
            \includegraphics[width=0.98\columnwidth]{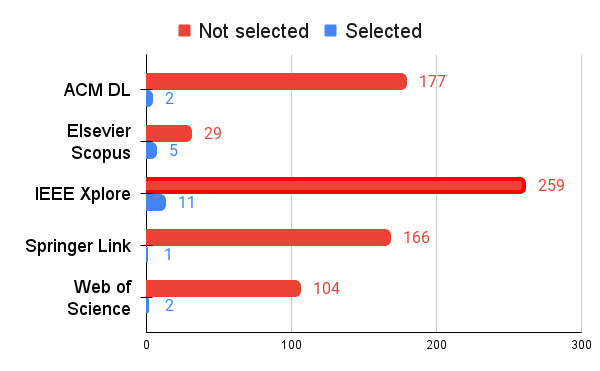}
            \caption{Selected and not selected studies in each of the search sources.}
            \label{fig:graficos_totais_busca_em_bases}
\end{figure}

We found 259 studies in IEEE Xplore, 166 in Springer Link, 104 in Web of Science, 177 in ACM Digital Library, and 29 in Elsevier Scopus search source. After reading the studies, we identified that 67 were duplicated, and 640 fitted some exclusion criteria. Therefore, 21 studies were selected to extract information. Figure~\ref{fig:graficos_totais_busca_em_bases} shows the number of studies found and the number of studies selected for information extraction.

\subsection{Description}~\label{subsec:sr-description}
        
The last step of the SR consists of reading all the selected studies to verify if it has information that can be extracted to answer the SQs defined in the planning step. In this section, we describe the selected studies.

Chermak and Aouf (2012)~\cite{chermak2012} study detecting and matching FPs in scenes with high-light and low-light areas using HDR images. The authors use an LDR capture system and an HDR capture system to compare the effectiveness of using HDR images. Using the software FLANN to match FPs, the authors reported that the matching improved up to $29.35$ times when using the HDR capture system.

Kontogianni et al. (2015)~\cite{kontogianni2015} evaluate FP detection in scenes containing buildings. The Mantiuk~\cite{tm_mantiuk2006} TM algorithm is used to convert HDR images into LDR (TM-LDR images), and FPs are described using SIFT, SURF, FAST, and ORB. The authors observed that the ORB algorithm detected more FPs when using the TM-LDR images. ORB also performed faster when compared to the other descriptors used.

Jagadish and Sinzinger (2008)~\cite{jagadish2008} use a detector focused on detecting junction points in the façade of buildings, i.e., corners and vertices~\cite{sinzingerDetector}. A rotation invariant descriptor~\cite{sinzingerDescriptor} and SIFT algorithm is used as the benchmark. The authors observed an increase of $19.35\%$ in the matching and a $15\%$ better sensitivity in the proposed approach compared to SIFT.

Zhuang and Liang (2019)~\cite{zhuang2019} propose an approach for extracting locally invariant FPs. The authors developed two TMs: one based on the reflection of light in the scene (called \textit{reflection layer}) and another based on the scene's lighting (called \textit{illumination layer}). The FP detection uses the reflection layer, while the description uses the illumination layer. For FP detection, the authors used the FAST detector and the binary descriptor for description. With this approach, the authors report that there was an improvement in FP detection when compared to using HDR images. While the approach using HDR images detected 11 FPs, 990 FPs were detected using the proposed approach. In FP matching, using an HDR image resulted in 4 FPs matched, while when using the proposed approach, 139  were matched.

Ige et al. (2016)~\cite{ige2016} used support vector machines (SVMs) and local binary pattern (LBP) for the facial expression recognition task. The proposed method receives image descriptions as input. For description, the authors used SURF with original LDR images and a TM-LDR image generated when applying a TM algorithm~\cite{tm_mantiuk2006} to the HDR image. The authors reported that the TM-LDR image approach presented better results, reaching 79.8\% accuracy. In contrast, the traditional methods---without using SVM, LBP, and using LDR images---obtained between 31.3 \% and 70.8\% accuracy.

Mukheerjee et al.~\cite{mukherjee2021} proposed a convolutional neural network (CNN) that receives an HDR image for object detection. They propose a methodology to generate and validate a large scale annotated HDR dataset from an existing LDR dataset, and create an out of distribution (OOD) HDR dataset to test and compare the performance of HDR and LDR trained detectors under extreme lighting conditions. To generate the HDR dataset, they use the Kovaleski et al.~\cite{kovaleski2014}; Huo et al.~\cite{huo2014}, Eilersten et al.~\cite{eilertsen2017}, and Marnerides et al.~\cite{marnerides2018} expansion operators (EOs) to expand LDR images into HDR. The FasterRCNN~\cite{fasterrcnn}, SSD300 and SSD512~\cite{ssd300} detectors and PascalVOC 2007~\cite{pascalVOC2007}, PascalVOC 2012~\cite{pascalVOC2012} and a personal dataset was used. Authors reported that using the proposed methodology, HDR trained models are able to achieve from 10 to 12\% more accuracy compared to LDR trained models.

Due to their continuity, four groups of studies were organized by author and theme: Section~\ref{subsubsec:tr-works_melo_nascimento} describes three works developed by Melo et al. and Nascimento et al.; Section~\ref{subsubsec:tr-works_prybil} describes two studies developed by P{\v{r}}ibyl et al.; Section~\ref{subsubsec:tr-works_rana} describes six studies developed by Rana et al.; and Section~\ref{subsubsec:tr-tracking_localization} describes studies focused on tracking and localization.

\subsubsection{Melo et al. and Nascimento et al. studies}~\label{subsubsec:tr-works_melo_nascimento}
            
Melo et al. (2018)~\cite{welerson_hdr} describe an approach using HDR images as input to detection algorithms. The SIFT Difference of Gaussian (DoG) detector and the Harris corner detector are modified to support HDR images by adding one more step in the detection with the application of a local mask based on the coefficient of variation (CV). In Harris corner, the CV mask is applied after the Gaussian filter, while in DoG, the CV mask is applied to all images of the scale space. The 2D and 3D lighting datasets from P{\v{r}}ibyl et al. (2012)~\cite{pvribyl2012} were used for testing. A significant improvement was obtained in the uniformity rate metric when using HDR images compared to their LDR versions.

Nascimento et al. (2022)~\cite{nascimento2022} adapted the canonical SIFT algorithm to receive HDR images and compared the description performance from the algorithms when using LDR and HDR images. Using mAP metric, 2D lighting and 3D lighting datasets from P{\v{r}}ibyl et al.~\cite{pvribyl2012}, results show that using HDR images increase mAP by 61.50\% in 2D lighting and 81.80\% in 3D lighting dataset.

Nascimento et al. (2023)~\cite{nascimento2023} proposes a new detector, based on coefficient of variation (CV) and designed for HDR images, called DetectorCV. The DetectorCV is compared with Harris~\cite{harris1988}, DoG~\cite{lowe2004}, SURF~\cite{surf}, Harris for HDR, DoG for HDR and SURF for HDR~\cite{welerson_hdr} detectors, using Uniiformity~\cite{welerson_hdr} and RR metrics, with ProjectRoom dataset from Rana et al.~\cite{rana2015}, 2D lighting and 3D lighting datasets from P{\v{r}}ibyl et al.~\cite{pvribyl2012}. Results show that DetectorCV outperforms other detectors in Uniformity metric. The DetectorCV outperforms Harris, DoG, DoG for HDR and SURF for HDR algorithms in RR metric.

\subsubsection{P{\v{r}}ibyl et al. studies}~\label{subsubsec:tr-works_prybil}

P{\v{r}}ibyl et al. (2012)~\cite{pvribyl2012} investigates if HDR imagery improves FP detection performance when compared to LDR images. The authors have elaborated a dataset~\footnote{Dataset by P{\v{r}}ibyl et al.~\cite{pvribyl2012}, available at: \url{http://www.fit.vutbr.cz/~ipribyl/FPinHDR/dataset\_JVCI}. (Accessed on July 01, 2023).} with equivalent scene captures in LDR and HDR images. The scenes were captured in LDR format, with 10 different exposure times. Then, Debevec algorithm~\cite{debevec2008} was used to generate HDR captures. The dataset consists of two scenes: a planar (2D) scene containing three posters in A4 sheets next to each other, attached to a box; and a 3D scene containing several non-planar rigid objects. Each scene is captured in three different sequences:

\begin{itemize}
    \item \textbf{Viewpoint}: the scene is captured 21 times in a circular trajectory around the scene's center with a step of $2.5$ degrees, resulting in a viewpoint range of 50 degrees.
    \item \textbf{Distance}: the scene is captured seven times, and the distance between the camera and the scene increases exponentially, yielding the distance sequence of 100, 103, 109, 122, 147, 197, and 297 cm.
    \item \textbf{Lighting}: the scene is captured seven times, each time with different combinations of 3 light sources being on or off, with at least one on.
\end{itemize}

\begin{figure*}[!htb]
    \centering
    \subfloat[\label{fig:dataset_prybil_pv1}]{%
        \includegraphics[width=0.15\linewidth]{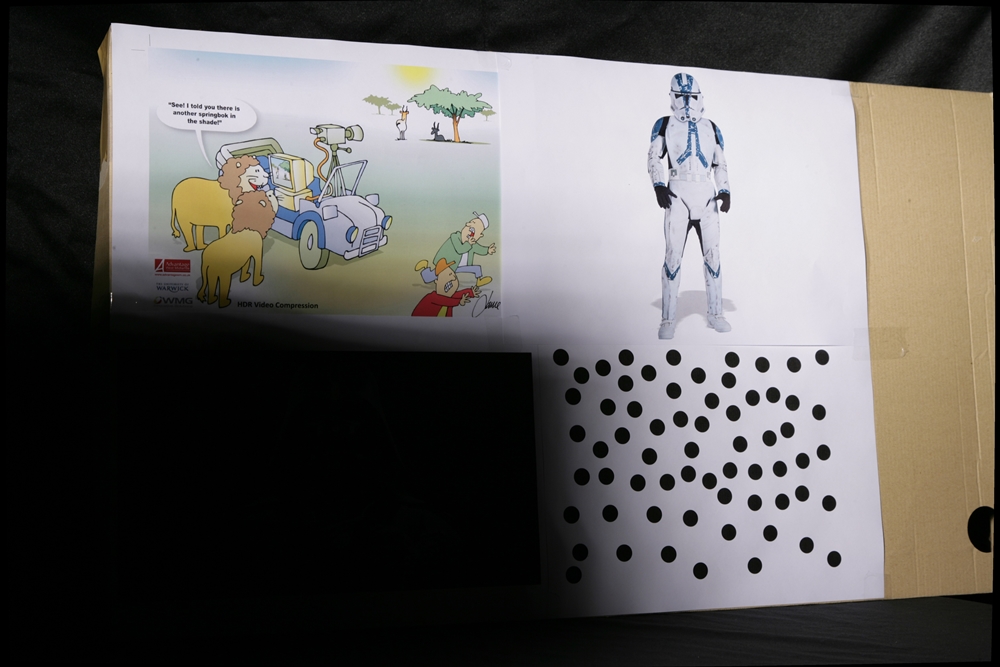}
    }
    \hfill
    \subfloat[\label{fig:dataset_prybil_pv3}]{%
        \includegraphics[width=0.15\linewidth]{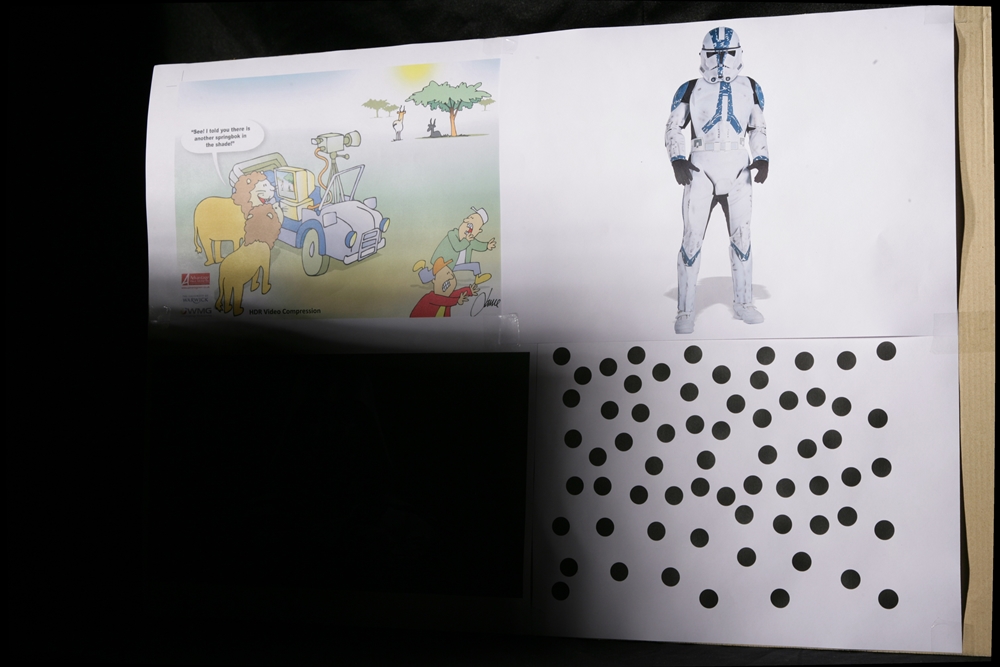}
    }
    \hfill
    \subfloat[\label{fig:dataset_prybil_ds1}]{%
        \includegraphics[width=0.15\linewidth]{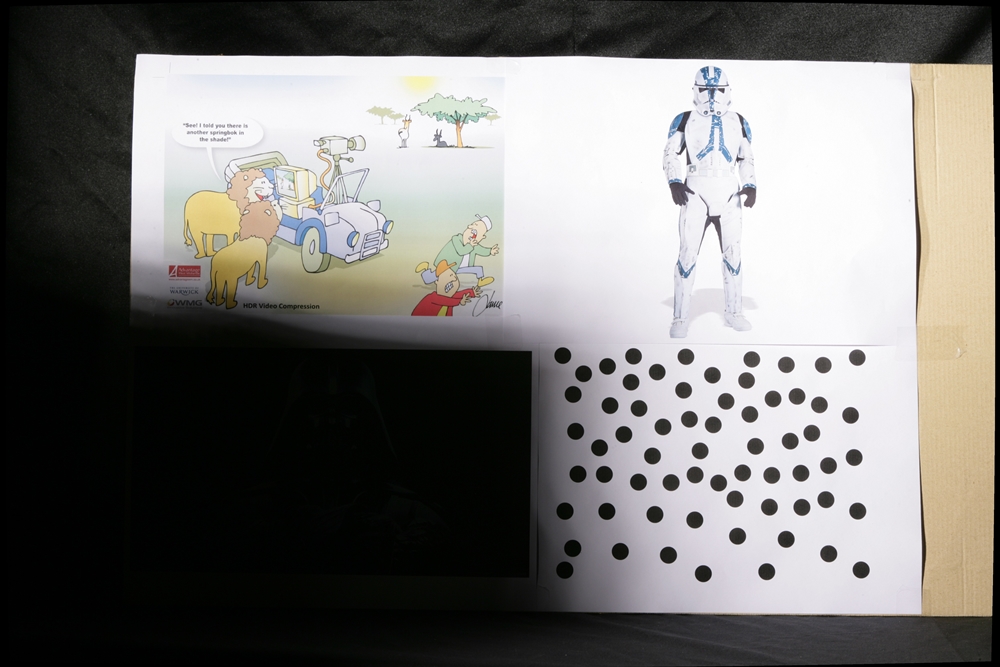}
    }
    \hfill
    \subfloat[\label{fig:dataset_prybil_ds3}]{%
        \includegraphics[width=0.15\linewidth]{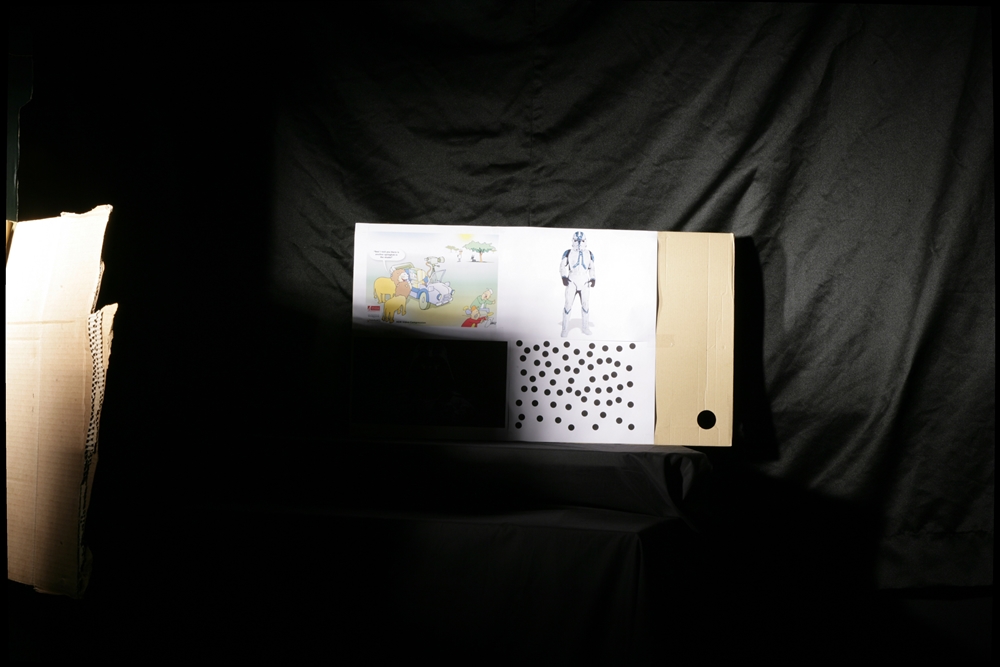}
    }
    \hfill
    \subfloat[\label{fig:dataset_prybil_il1}]{%
        \includegraphics[width=0.15\linewidth]{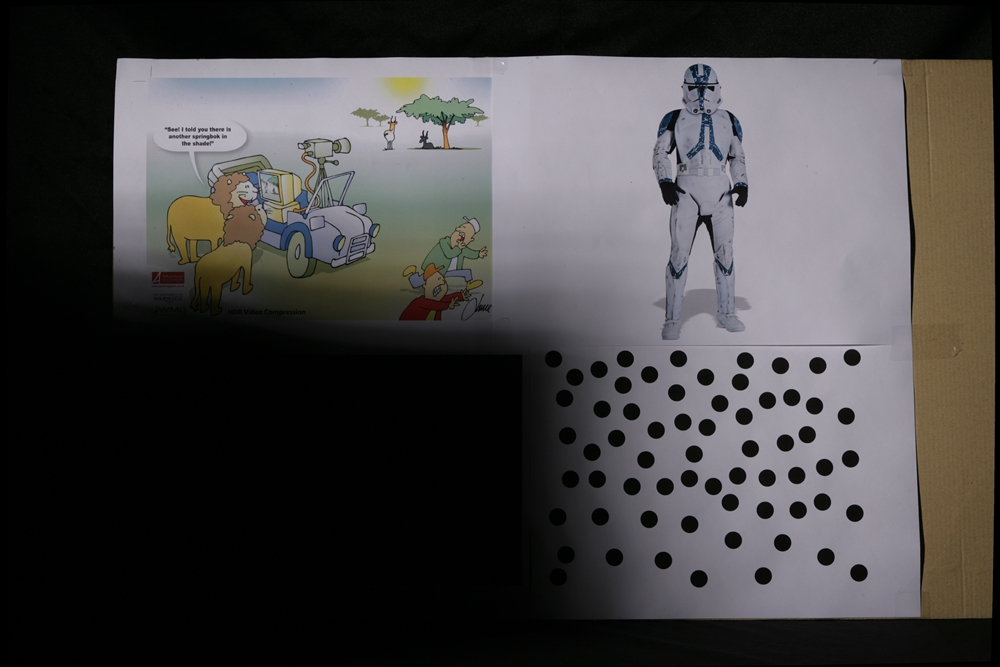}
    }
    \hfill
    \subfloat[\label{fig:dataset_prybil_il3}]{%
        \includegraphics[width=0.15\linewidth]{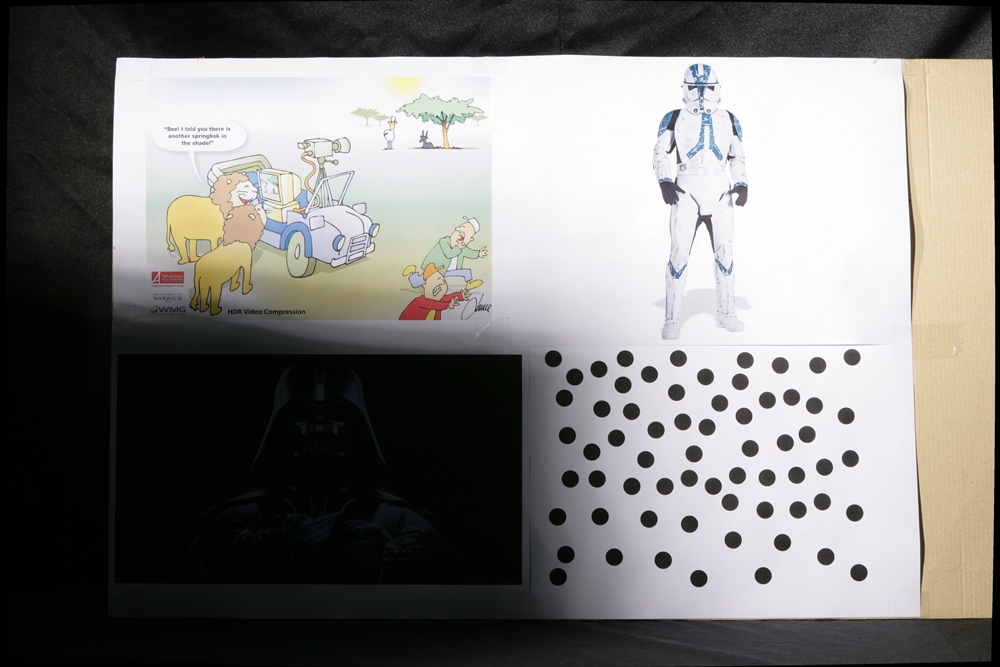}
    }
    \caption{Capture sequences examples from P{\v{r}}ibyl et al. (2012)~\cite{pvribyl2012} 2D dataset. (a) and (b) from viewpoint capture sequence, (c) and (d) from distance capture sequence, and (e) and (f) from illumination capture sequence.}
    \label{fig:dataset_prybil_2d_il} 
\end{figure*}

The authors used a $\log_2$ transformation as global tone mapping (GTM) and Zuderveld~\cite{tm_zuiderveld1994} algorithm as local tone mapping (LTM) to transform the HDR images into new tone-mapped LDR versions of the captures. The authors added preprocessed LDR images using Wallis-Filter~\cite{wallis_filter} (WAL) to equalize contrast. Figure~\ref{fig:dataset_prybil_2d_il} shows examples of the capture sequences from P{\v{r}}ibyl et al. (2012) dataset~\cite{pvribyl2012}.
    
P{\v{r}}ibyl et al. (2012)~\cite{pvribyl2012} compared Harris corner~\cite{harris1988}, Shi-Tomasi~\cite{shi-tomasi}, FAST~\cite{fast} and SURF~\cite{surf} detectors performance using the LDR and tone-mapped LDR (TM-LDR) versions of the captures. As a result, they found that the approaches using GTM and LTM resulted in a higher repeatability rate (RR) than the others. They also observed that, on average, the 2D dataset had a lower RR than the 3D dataset.
    
P{\v{r}}ibyl et al. (2016)~\cite{pvribyl2016} used several TM algorithms to transform HDR into LDR images, used the resulting TM-LDR images as input to FP detection algorithms, and compared the FP detection when using the different TM-LDR images. Using the dataset elaborated in their previous study~\cite{pvribyl2012}, they generated six different versions of the captures using global TMs, five using local TMs, one LDR capture, one LDR preprocessed version using canonical histogram equalization, one LDR preprocessed version using CLAHE histogram equalization~\cite{clahe}, one version using a linear mapping from HDR to LDR, and one version using a logarithmic mapping from HDR to LDR. They used Harris corner~\cite{harris1988}, Shi-Tomasi~\cite{shi-tomasi}, FAST~\cite{fast}, SURF~\cite{surf} and SIFT (DoG)~\cite{lowe2004} as detectors.
    
P{\v{r}}ibyl et al. (2016)~\cite{pvribyl2016} observed that global TMs have the side effect of compressing the contrast of the intermediate tones of the image since they take into account the global illumination. The values of the very light and very dark areas have greater weight than the intermediate values. Among the TMs used, Reinhard et al.~\cite{tm_reinhard2005} present the worst performance, detecting few FPs. The FAST detector presented poor performance when used with global TMs, being sensitive to contrast compression. 
            
On the other hand, when using local TM, more FPs were detected compared to when global TMs were used. Even so, detection in dark areas proved difficult, given that few or no FPs were detected. When using Fattal~\cite{tm_fattal2002} and Mantiuk~\cite{tm_mantiuk2006} TM algorithms, a higher number of FPs were detected when compared to other TM algorithms~\cite{pvribyl2016}.

\subsubsection{Rana et al. studies}~\label{subsubsec:tr-works_rana}

\begin{figure}[!htb]
    \centering
    \subfloat[PR 1.\label{fig:dataset_rana_pr1}]{%
        \includegraphics[width=0.3\linewidth]{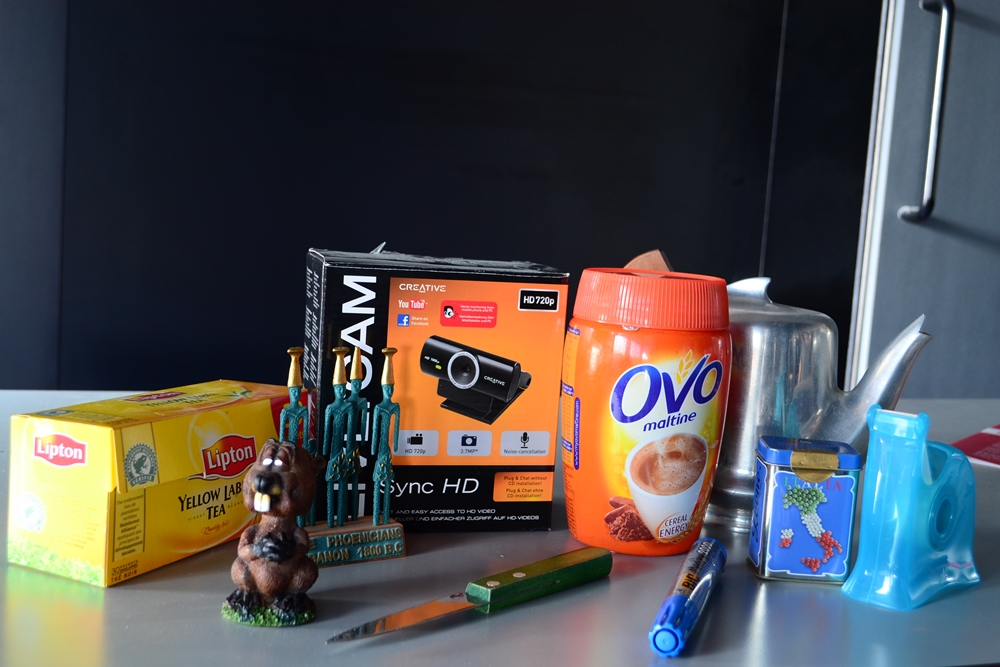}
    }
    \hfill
    \subfloat[PR 2.\label{fig:dataset_rana_pr2}]{%
        \includegraphics[width=0.3\linewidth]{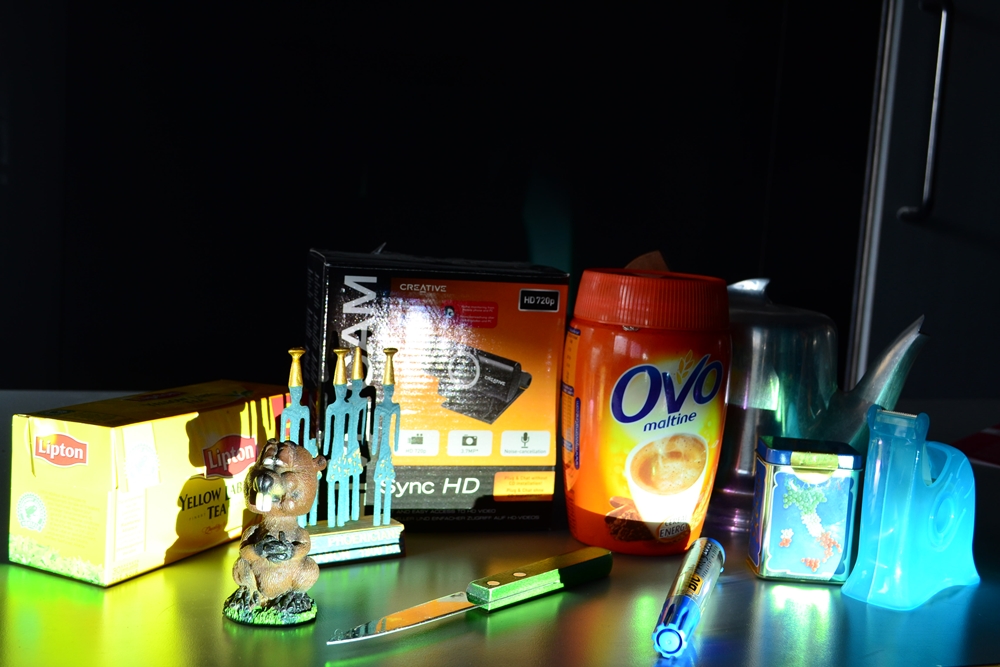}
    }
    \hfill
    \subfloat[PR 3.\label{fig:dataset_rana_pr3}]{%
        \includegraphics[width=0.3\linewidth]{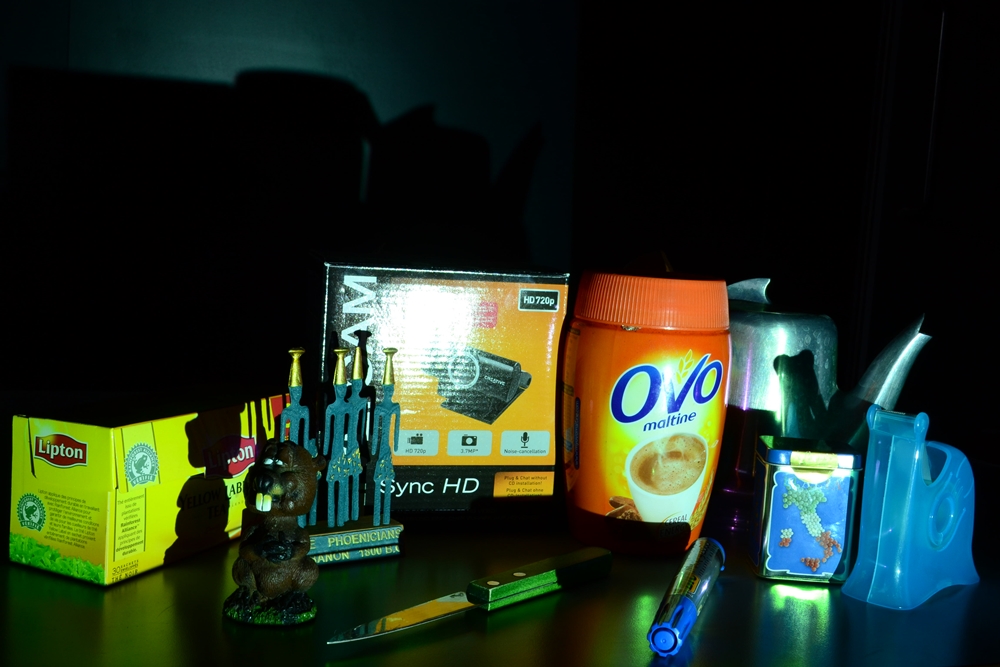}
    }
    \hfill
    \subfloat[LR 1\label{fig:dataset_rana_lr1}]{%
        \includegraphics[width=0.3\linewidth]{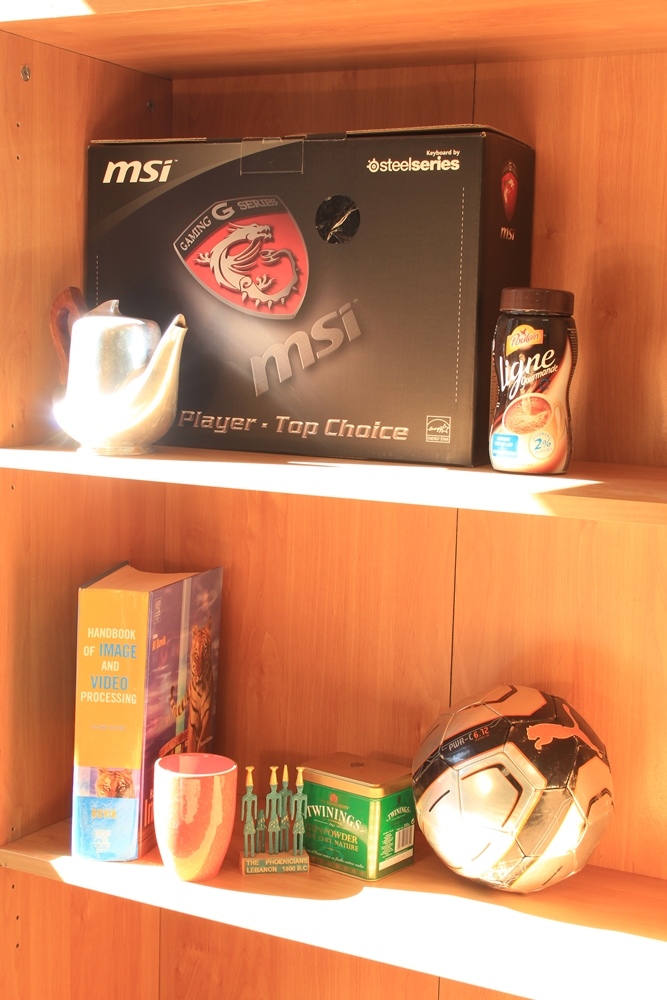}
    }
    \hfill
    \subfloat[LR 2.\label{fig:dataset_rana_lr2}]{%
        \includegraphics[width=0.3\linewidth]{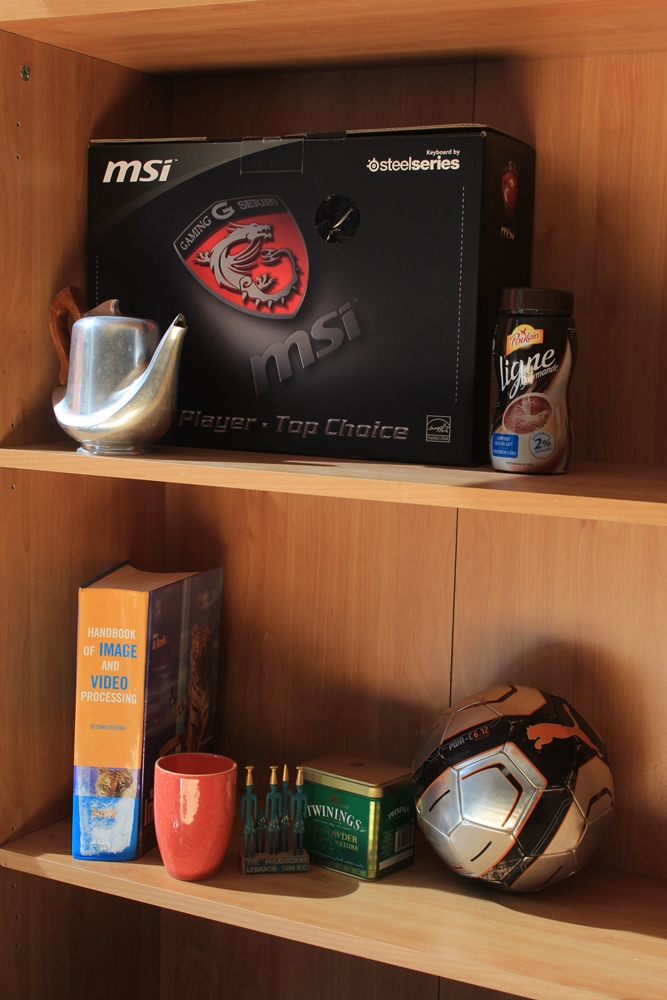}
    }
    \hfill
    \subfloat[LR 3.\label{fig:dataset_rana_lr3}]{%
        \includegraphics[width=0.3\linewidth]{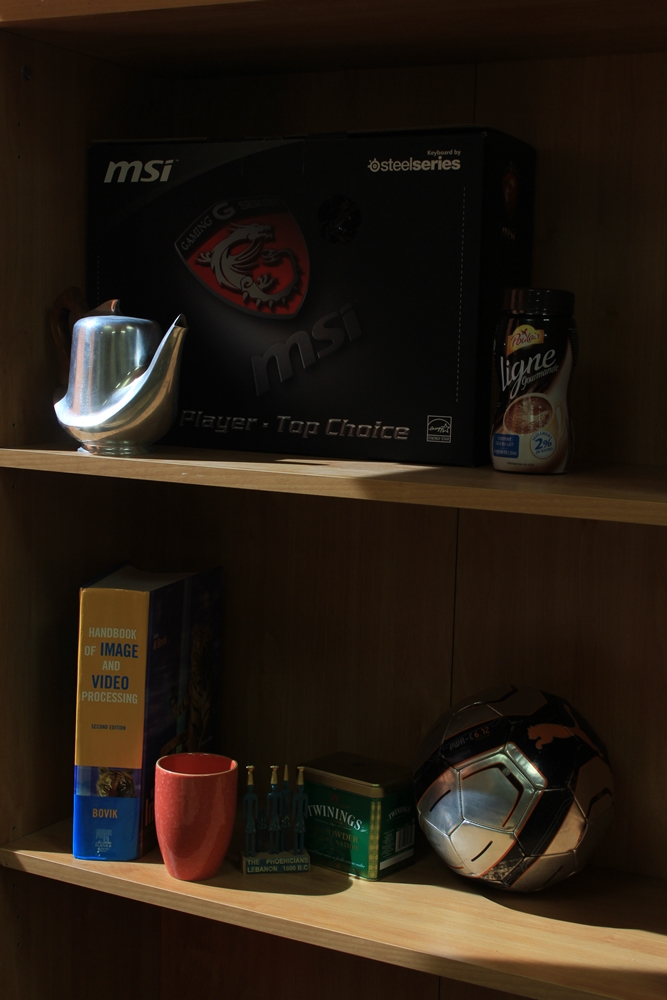}
    }
    \caption{Capture examples from Rana et al. (2015)~\cite{rana2015} dataset. (a), (b) and (c) from ProjectRoom scene, and (d), (e), and (f) from LightRoom scene.}
    \label{fig:dataset_rana} 
\end{figure}

Rana et al. investigates the use TM algorithms to convert HDR into LDR images and use the TM-LDR images as input to detection and description algorithms. First, Rana et al.~\cite{rana2015} elaborated a dataset~\footnote{Dataset by Rana et al.~\cite{rana2015}, available at: \url{http://webpages.l2s.centralesupelec.fr/perso/giuseppe. valenzise/sw/HDR\%20Scenes.zip}. (Accessed on July 01, 2023).} of two scenes: ProjectRoom (PR) and LightRoom (LR). The PR scene comprises eight different lighting configurations created by blocking a subset of the light sources. The LR dataset comprises seven different natural lighting conditions obtained by changing how closed the room blinds are and the position of a tungsten lamp~\cite{rana2015}. Similarly to P{\v{r}}ibyl et al.~\cite{pvribyl2012}, Rana captured the dataset using LDR techniques and generated the HDR images using Debevec algorithm~\cite{debevec2008}. Figure~\ref{fig:dataset_rana} shows examples of captures from Rana et al.~\cite{rana2015} dataset.

In addition to the dataset, Rana et al. (2015)~\cite{rana2015} explored TM algorithms and used TM-LDR images as input to FP detection algorithms. The authors used different TMs than previous studies~\cite{pvribyl2012, pvribyl2016}. They used two global TMs, seven local TMs, linear mapping, logarithmic mapping, perceptually-uniform mapping~\cite{perceptual} (PU-HDR), with the 2D illumination and 3D illumination datasets from P{\v{r}}ibyl et al. (2012)~\cite{pvribyl2012}. When using SURF and Harris corner detectors, the authors observed that all approaches using TM images as input resulted in a better RR than when using LDR. The PU-HDR TM method showed better RR than global and local TMs.

Rana et al. (2016a)~\cite{rana2016a} studied the use of TM-LDR images to improve the performance of FP description algorithms. The authors used image matching, mean average precision (mAP), and mean repeatability rate (mRR) as performance metrics. SIFT~\cite{lowe2004}, SURF~\cite{surf}, FREAK~\cite{alahi2012freak}, and BRISK~\cite{leutenegger2011brisk} were used as description algorithms. The input images used were one LDR, two TM-LDR using global TMs, five TM-LDR using local TMs, one using a logarithmic mapping from HDR to LDR, and one linear mapping from HDR to LDR. The authors observed that the worst results were obtained using linear mapping images, while TM-LDR images presented better RR and mAP values. All TM-LDR techniques performed well with description algorithms, with marginal gains for SIFT.

Rana et al. (2016b)~\cite{rana2016b} investigated the optimization of retinex-based models using correlation coefficient (CC) and repeatability rate (RR) of reflectance images obtained from two retinex-based TMs: bilateral tone mapping operator (BTMO) and Gaussian tone mapping operator (GTMO). Four algorithms were created: BTMO optimized with CC (CCBTM), BTMO optimized with RR (RRBTM), GTMO optimized with CC (CCGTM), and GTMO optimized with RR (RRGTM). Shi-Tomasi~\cite{shi-tomasi} and Harris corner~\cite{harris1988} were used as detection algorithms. In addition to the TM algorithms created, other local and global TM algorithms were used for comparison. The RR and CC metrics improved when using CCBTM, RRBTM, CCGTM, and RRGTM. The RRBTM and RRGTM detected more FPs~\cite{rana2016b}.

Rana et al. (2017a)~\cite{rana2017a} developed the learning-based adaptive tone mapping operator (AdTMO), a framework to enhance FP detection stability under drastic illumination variations. The AdTMO is modulated based on a support vector regression (SVR) model using local higher-order characteristics. To compare with AdTMO, the authors used the GTMO, BTMO~\cite{rana2016b}, Chiu~\cite{tm_chiu1996}, Drago~\cite{tm_drago2003}, Reinhard~\cite{tm_reinhard2005}, and Mantiuk~\cite{tm_mantiuk2006} TM algorithms. The Harris corner~\cite{harris1988}, Shi-Tomasi~\cite{shi-tomasi}, FAST~\cite{fast}, SURF~\cite{surf}, and SIFT~\cite{lowe2004} were used as detectors. The authors showed that the AdTMO presents a significantly better RR metric in all cases, with rates between 10\% and 45\% higher than the second best operator (BTMO)~\cite{rana2017a}.

Rana et al. (2017b)~\cite{rana2017b} developed the learning-based descriptor-optimal tone mapping operator (DesTMO), a framework to tone-map HDR content for image matching under drastic illumination variations. A luminance invariant guidance model using an SVR to adapt the tone mapping function for image matching in scenes subject to a wide variety of illumination changes was used. The authors observed that DesTMO had significantly better rates of correct FP matches and mAP than the other TMs used in the comparison~\cite{rana2017b}.
            
Finally, Rana et al. (2019)~\cite{rana2019} developed a two-step framework, called (OpTMO), composed of a luminance-invariant guidance model based on an SVR to optimally adapt the tone mapping function for image matching and an energy maximization model to generate appropriate training samples for learning the SVR. At each step, the framework collectively addresses both FP detection and description steps in the feature matching. OpTMO is compared with some conventional TMs, AdTMO~\cite{rana2016b}, and DesTMO~\cite{rana2017b}. The experimental results show that OpTMO is equivalent to DesTMO in FPs detection, presenting similar repeatability. As for its use in image matching, OpTMO presented better RR than all the others TMs, with rates between 10\% and 53\% better than the second best-placed, DesTMO~\cite{rana2019}.        

\subsubsection{Tracking and localization studies}~\label{subsubsec:tr-tracking_localization}
            
Yeh et al. (2021)~\cite{yeh2021} evaluates the use of HDR-based images to run an RGB-D SLAM~\cite{rgbd_slam} framework. The authors used a deep learning model that takes an LDR image as input and generates an HDR image, that is used to generate a radiance map which is normalized. An ORB descriptor is trained to receive normalized radiance maps and generate its description. The dataset TUM RGB-D~\cite{sturm2012benchmark} and a private dataset are used to evaluate the deep learning model. Yeh et al. (2021) used image matching as metric, comparing the results obtained using the ORB-SLAM2~\cite{mur2017orb} algorithm, the proposed framework with normalized radiance map, and the proposed framework with radiance map not normalized to verify if the radiance map normalization influences the image matching. The authors observed that the framework using the normalized radiance map presented a higher mean detection than the other approaches.

Albrecht et al. (2019)~\cite{alberecht2019} evaluated the usage of TM algorithms to transform HDR into LDR images and use the resulting TM-LDR images as input to the ORB-SLAM2 algorithm. The HDR images were generated using Mertens et al.~\cite{mertensHDR} algorithm. With HDR images, some TMs were applied and used as input to an ORB-SLAM2 algorithm. The authors observed that using TM-LDR images brought benefits in complex lighting scenes. However, the tracking algorithm fails in critical situations such as lighting transition (leaving a dimly lit room to an open sunny environment, for example).

Chermak et al. (2014)~\cite{chermak2014} used LDR and HDR images as input to the Kanade–Lucas–Tomasi (KLT) tracking algorithm, as implemented in OpenCV. An LDR capture system and an HDR capture system were used with SIFT~\cite{lowe2004}, SURF~\cite{surf}, GFTT~\cite{shi-tomasi} and FAST~\cite{fast} algorithms. The authors observed that the approach using HDR images showed results up to 29.35 times better when compared to those using LDR images.

Jinno et al. (2013)~\cite{jinno2013} developed a TM algorithm specialized in security systems that capture HDR images to improve object tracking in the scene. The authors modified the SIFT in order to optimize the processing time. The modification consists of using a mask to remove the image's background, reducing the area to be processed. The authors observed that, compared to other TMs, the developed method significantly improves object tracking. 

\subsubsection{Systematic review conclusions}~\label{subsec:tr-sr_conclusion}
            
Using HDR images as input to detector and descriptor algorithms requires changing these algorithms to support the dynamic range of HDR images correctly. Because of this necessity, most studies use a TM algorithm to enhance the image and get better results in detection and description. Some studies have proposed new TMs to improve specific tasks, such as invariance to environment illumination~\cite{rana2017b} or object tracking in security cameras~\cite{jinno2013}.
            
Mantiuk, Reinhard, Drago, and Fattal were the most frequently used TM algorithms, with respectively 9, 8, 6, and 6 studies using them. The most used FP extraction algorithms are SURF~\cite{surf} (11 studies), SIFT\cite{lowe2004} (10 studies), Harris~\cite{harris1988} (8 studies), and FAST (8 studies).
            
Yeh et al.~\cite{yeh2021} used an HDR image to calculate a normalized radiance map and used it as input to their proposed framework. On the other hand, only Melo et al. (2018)~\cite{welerson_hdr} and Nascimento et al. (2022)~\cite{nascimento2022} used HDR images as input to detection and description algorithms. Furthermore, only two studies proposed and made their datasets available~\cite{pvribyl2012, rana2015}.
            

\begin{table*}
\centering
\rotatebox{90}{
\begin{minipage}{1.25\textwidth}
\caption{Systematic review summary.}
\label{tab:rs_dados}
\resizebox{\textwidth}{!}{%
\begin{tabular}{
    >{\columncolor[HTML]{EFEFEF}}c |c|c|c|c|c|c|c|}
    \cline{2-8}
    \cellcolor[HTML]{FFFFFF}{\color[HTML]{333333} } & \cellcolor[HTML]{EFEFEF}{\color[HTML]{330001} \textbf{Use HDR?}} & \cellcolor[HTML]{EFEFEF}{\color[HTML]{330001} \textbf{TM algorithms}} & \cellcolor[HTML]{EFEFEF}{\color[HTML]{330001} \textbf{Detector}} & \cellcolor[HTML]{EFEFEF}{\color[HTML]{330001} \textbf{Detector metrics}} & \cellcolor[HTML]{EFEFEF}{\color[HTML]{330001} \textbf{Descriptor}} & \cellcolor[HTML]{EFEFEF}{\color[HTML]{330001} \textbf{Descriptor metrics}} & \cellcolor[HTML]{EFEFEF}{\color[HTML]{330001} \textbf{Used dataset}} \\ \hline

    \multicolumn{1}{|c|}{\cellcolor[HTML]{EFEFEF}{\color[HTML]{333333} P{\v{r}}ibyl et al. (2012)~\cite{pvribyl2012}}}     & --  & \cite{tm_zuiderveld1994} & \makecell{Harris Corner, GFTT,\\FAST, SURF} & RR & -- & -- & \cite{pvribyl2012} \\ \hline
    \multicolumn{1}{|c|}{\cellcolor[HTML]{EFEFEF}{\color[HTML]{333333} Chermak et al. (2012)~\cite{chermak2012}}}     & Yes & -- & \makecell{SIFT, Harris,\\GFTT, SURF,\\FAST} & -- & SIFT, SURF & FLANN accuracy & \makecell{personal,\\not available} \\ \hline
    \multicolumn{1}{|c|}{\cellcolor[HTML]{EFEFEF}{\color[HTML]{333333} Jinno et al. (2013)~\cite{jinno2013}}}       & --  & \cite{tm_reinhard2002} , \cite{tm_li2005} & Modified SIFT & -- & Modified SIFT & \makecell{Tracking\\accuracy} & \makecell{personal,\\not available} \\ \hline
    \multicolumn{1}{|c|}{\cellcolor[HTML]{EFEFEF}{\color[HTML]{333333} Chermak et al.(2014)~\cite{chermak2014}}}     & Yes & -- & \makecell{SIFT, SURF,\\GFTT, FAST} & Amount of FPs & KLT (tracking) & tracking performance & \makecell{personal,\\not available} \\ \hline
    \multicolumn{1}{|c|}{\cellcolor[HTML]{EFEFEF}{\color[HTML]{333333} Rana et al. (2015)~\cite{rana2015}}}          & --  & \makecell{\cite{tm_chiu1996}, \cite{tm_drago2003}  ,\\\cite{tm_fattal2002}, \cite{tm_larson1997},\\\cite{tm_ashikmin2002}, \cite{tm_mantiuk2006}  ,\\\cite{tm_pattanaik2002}, \cite{tm_reinhard2002} ,\\\cite{tm_schlick1994}} & Harris, SURF & mRR & -- & -- & \cite{rana2015} \\ \hline
    \multicolumn{1}{|c|}{\cellcolor[HTML]{EFEFEF}{\color[HTML]{333333} Kontogianni et al. (2015)~\cite{kontogianni2015}}} & --  & \cite{tm_mantiuk2006}   & \makecell{SIFT, SURF,\\FAST and ORB} & \makecell{Amount of detected FPs\\and processing time} & -- & -- & \makecell{personal,\\not available} \\ \hline
    \multicolumn{1}{|c|}{\cellcolor[HTML]{EFEFEF}{\color[HTML]{333333} P{\v{r}}ibyl et al. (2016)~\cite{pvribyl2016}}}     & --  & \makecell{\cite{tm_mantiuk2006}  , \cite{tm_larson1997},\\\cite{tm_reinhard2002} , \cite{tm_reinhard2005},\\\cite{tm_mantiuk2008}, \cite{tm_fattal2002},\\\cite{tm_fattal2009}, \cite{tm_kiser2012},\\\cite{tm_yates2008}} & \makecell{Harris, GFTT,\\SIFT, SURF, FAST,\\BRISK} & mRR & -- & -- & \cite{pvribyl2012} \\ \hline
    \multicolumn{1}{|c|}{\cellcolor[HTML]{EFEFEF}{\color[HTML]{333333} Ige et al. (2016)~\cite{ige2016}}}         & --  & \cite{tm_mantiuk2006}   & LBP & -- & SURF & \makecell{Face recognition\\accuracy} & \makecell{personal,\\not available} \\ \hline
    \multicolumn{1}{|c|}{\cellcolor[HTML]{EFEFEF}{\color[HTML]{333333} Rana et al. (2016a)~\cite{rana2016a}}}          & --  & \makecell{\cite{tm_reinhard2002} , \cite{tm_drago2003}  ,\\\cite{tm_mantiuk2006}  , \cite{tm_fattal2002},\\\cite{tm_chiu1996}, \cite{tm_durand2002}} & \makecell{SIFT, SURF, \\FREAK, BRISK} & mRR & \makecell{SIFT, SURF, \\FREAK, BRISK} & mAP & \cite{rana2015} \\ \hline
    \multicolumn{1}{|c|}{\cellcolor[HTML]{EFEFEF}{\color[HTML]{333333} Rana et al. (2016b)~\cite{rana2016b}}}          & --  & \makecell{\cite{tm_mantiuk2006}  , \cite{tm_reinhard2002} ,\\\cite{tm_chiu1996}, \cite{tm_durand2002},\\\cite{tm_drago2003}  } & Harris and SURF & RR and CC & -- & -- & \cite{rana2015} \\ \hline
    \multicolumn{1}{|c|}{\cellcolor[HTML]{EFEFEF}{\color[HTML]{333333} Rana et al. (2017a)~\cite{rana2017a}}}          & --  & \makecell{\cite{tm_reinhard2002} , \cite{tm_mantiuk2006}  ,\\\cite{rana2016b}, \cite{tm_drago2003}  ,\\\cite{tm_chiu1996}, \cite{tm_durand2002}} & \makecell{Harris, GFTT,\\FAST, BRISK,\\SURF and SIFT} & RR & -- & -- & \cite{rana2015} \\ \hline
    \multicolumn{1}{|c|}{\cellcolor[HTML]{EFEFEF}{\color[HTML]{333333} Rana et al. (2017b)~\cite{rana2017b}}}          & --  & \makecell{\cite{tm_reinhard2002} , \cite{tm_mantiuk2006}  ,\\\cite{tm_chiu1996}, \cite{rana2016b},\\\cite{tm_drago2003}  } & \makecell{SURF, SIFT, \\FREAK, BRISK} & -- & \makecell{SURF, SIFT,\\FREAK, BRISK} & mAP and matching score & \cite{rana2015} \\ \hline
    \multicolumn{1}{|c|}{\cellcolor[HTML]{EFEFEF}{\color[HTML]{333333} Jagadish et al. (2018)~\cite{jagadish2008}}}    & --  & \cite{tm_fattal2002} & \cite{sinzingerDetector} & -- & \cite{sinzingerDetector} & Amount & \makecell{personal,\\not available} \\ \hline
    \multicolumn{1}{|c|}{\cellcolor[HTML]{EFEFEF}{\color[HTML]{333333} Melo et al. (2018)~\cite{welerson_hdr}}}    & Yes & \cite{tm_fattal2002}, \cite{tm_mantiuk2006}   & Harris, SIFT & mRR, uniformity rate & -- & -- & \cite{pvribyl2012}, \cite{rana2015} \\ \hline
    \multicolumn{1}{|c|}{\cellcolor[HTML]{EFEFEF}{\color[HTML]{333333} Zhuang et al. (2019)~\cite{zhuang2019}}}      & Yes & -- & BRISK & Amount of FPs & BRISK & \makecell{amount of\\matched FPs} & \makecell{personal,\\not available} \\ \hline    
    \multicolumn{1}{|c|}{\cellcolor[HTML]{EFEFEF}{\color[HTML]{333333} Alberecht et al. (2019)~\cite{alberecht2019}}}   & --  & \makecell{\cite{tm_durand2002}, \cite{tm_mantiuk2006}  ,\\\cite{tm_drago2003}  , \cite{tm_reinhard2010}} & ORB & -- & ORB & Analisys with SLAM & \makecell{personal,\\not available} \\ \hline
    \multicolumn{1}{|c|}{\cellcolor[HTML]{EFEFEF}{\color[HTML]{333333} Rana et al. (2019)~\cite{rana2019}}}          & --  & \makecell{\cite{tm_drago2003}  , \cite{tm_chiu1996},\\\cite{tm_durand2002}, \cite{tm_reinhard2002} , \cite{tm_chalmers2017},\\\cite{tm_ledda2005}, \cite{tm_cadik2008}} & \makecell{Harris, BRISK,\\FREAK, FAST,\\SURF and SIFT} & RR & \makecell{BRISK, FREAK,\\SURF and SIFT} & mAP & \cite{rana2015} \\ \hline
    \multicolumn{1}{|c|}{\cellcolor[HTML]{EFEFEF}{\color[HTML]{333333} Yeh et al. (2021)~\cite{yeh2021}}}         & Yes & -- & ORB & average detection & ORB & average matching & \makecell{personal, not available\\and TUM RGB-D} \\ \hline
    \multicolumn{1}{|c|}{\cellcolor[HTML]{EFEFEF}{\color[HTML]{333333} Mukherjee et al. (2021)~\cite{mukherjee2021}}}  & Yes & \cite{kovaleski2014}, \cite{huo2014}, \cite{eilertsen2017}, \cite{marnerides2018},  & \makecell{FasterRCNN~\cite{fasterrcnn},\\SSD300 and SSD512~\cite{ssd300}} & -- & ORB & mAP & \makecell{personal, not available\\\cite{pascalVOC2007}, and \cite{pascalVOC2012}} \\ \hline
    \multicolumn{1}{|c|}{\cellcolor[HTML]{EFEFEF}{\color[HTML]{333333} Nascimento et al. (2022)~\cite{nascimento2022}}}  & Yes & -- & DoG & -- & SIFT & mAP & \cite{pvribyl2012}, \cite{rana2015}  \\ \hline
    \multicolumn{1}{|c|}{\cellcolor[HTML]{EFEFEF}{\color[HTML]{333333} Nascimento et al. (2023)~\cite{nascimento2023}}}  & Yes & -- & \makecell{Harris, DoG, SURF,\\DetectorCV, Harris for HDR,\\SIFT for HDR, SURF for HDR} & RR, Uniformity & -- & -- & \cite{pvribyl2012}, \cite{rana2015}  \\ \hline

\end{tabular}%
}
\end{minipage}
}
\end{table*}
            
Table~\ref{tab:rs_dados} lists the TM methods, the detectors, the descriptors, the evaluation metrics, and the datasets used in each study.

\section{Methods}~\label{sec:methods}

In this study, we propose a library called CP\_HDR, capable of detection and description using LDR or HDR images as input. We compared the results obtained using LDR and HDR images. The library has state-of-the-art metrics (Section~\ref{subsec:metrics}), well-known detector and description algorithms with support to LDR and HDR images, modifications to improve detection and description in HDR images (Section~\ref{subsec:fp-algorithms}), and an automatic script to generate the intensity segmentation (Section~\ref{subsec:ROIs}). The experiment pipeline is explained in Section~\ref{subsec:pipeline}.

\subsection{Metrics}~\label{subsec:metrics}
        
As seen in Section~\ref{subsec:tr-sr_conclusion}, repeatability rate (RR) is the most used metric to evaluate FP detection. In the FP description, matching score and average precision (AP) are the most used metrics. Melo et al.~\cite{welerson_hdr} proposed the uniformity rate (UR) as an FP detection metric to verify whether an algorithm detects FPs in areas of the image with extreme lighting conditions. This metric is specially suitable to the purpose of this study, as we are comparing LDR and HDR detection and description algorithms. HDR images have a greater dynamic range than LDR images, so the UR can be used to verify how this feature influences the detection results in image areas with low or high luminosity.

\subsubsection{Repeatability rate (RR)}~\label{sec:met-repeatability}

The RR metric indicates the percentage of FPs in a reference image ($I_r$) that are detected in a test image ($I_t$). The RR evaluates the stability of the detector between captures of the same scene, resulting in a value in the interval $[0.0, 1.0]$. Values closer to 1 indicate that more FPs were detected in the two images. 

Let $n_r$ and $n_t$ be the total of FPs found in $I_r$ and $I_t$ respectively. Let $M$ be the maximum number of FPs considered in the analysis, and let $R_{rt}$ be the number of FPs found in both $I_r$ and $I_t$. The RR between $I_r$ and $I_t$ is defined by

\begin{equation}
    \operatorname{RR}(I_r, I_t) = \frac{R_{rt}}{\min(n_r, n_t, M)}
    \label{eq:rr}
\end{equation}
            
When there is no reference image but several captures of the same scene, we can use the summarized repeatability rate (sRR) as follows: we calculate the RR for all permutations between the captures of the scene and calculate the results average. All possible non-ordered pairs of different images are considered. Let $n$ be a number of captures in a given dataset. The sRR is defined by 
    
\begin{equation}
    \operatorname{sRR} = \frac{\sum_{i=1}^{n-1} \sum_{j=i+1}^{n} \operatorname{RR}(i, j)}{\sum_{k = 1}^{n - 1} k}
    \label{eq:rr2}
\end{equation}

\subsubsection{Uniformity rate (UR)}~\label{sec:met-metricas_uniformidade}            
The UR is a metric to measure the quality of FP detection in areas of the image with different luminosities~\cite{welerson_hdr}. In this study, we use the UR to evaluate how well the algorithms detect FPs in brightest, intermediate, and darkest areas of the image. We made an intensity segmentation of the image pixels, as detailed in Section~\ref{subsec:ROIs}. The value returned by the UR is in the interval $[0.0, 1.0]$, where the best result is $1.0$. The best result means that the FPs are equally distributed among groups of pixels with different illumination levels in the image. 
            
Let $T$ be the total number of detected FPs, let $n$ be the number of pre-divided groups of the image, and let $a_i$ be the number of FPs detected in the $i$-th group of the image. The UR is defined by

\begin{equation}
    \operatorname{U} = 1 - \left( \max_{1 \leq i \leq n} \left( \frac{a_i}{T} \right) - \min_{1 \leq i \leq n} \left( \frac{a_i}{T}\right) \right)
    \label{eq:u}
\end{equation} 

\subsubsection{Matching}~\label{sec:met-metricas_nndr}
            
To match FPs, we use the nearest neighbor distance ratio (NNDR)~\cite{mikolajczyk2005performance}. This quotient is often used to evaluate matching between two FPs~\cite{szeliski2010computerVision}. NNDR calculates the distance from the description vector to its nearest neighbor and its second nearest neighbor.

Let $d_1$ and $d_2$ be the distances to the nearest neighbor and the second nearest neighbor respectively. Let $D_t$ be the target description vector. Let $D_1$ and $D_2$ be the description vectors of the two nearest neighbors of $D_t$. NNDR is defined by 

\begin{equation}
    \operatorname{NNDR} = \frac{d_1}{d_2} = \frac{\left \| D_t - D_1 \right \|}{\left \| D_t - D_2 \right \|} < \operatorname{th}
    \label{eq:nndr}
\end{equation}
            
The matching between two description vectors is considered as true positive (tp) if the NNDR value is less than a defined threshold ($\operatorname{th}$) and false positive (fp) otherwise. We used $0.7$ as the threshold value and the Euclidean distance to calculate the distance between two description vectors~\cite{nascimento2022}.

\subsubsection{Mean average precision (mAP)}~\label{sec:met-metricas_mAP}

The average precision (AP) metric evaluates matching between FPs obtained from two captures of the same scene. The AP can be obtained by calculating the area under the ROC curve~\cite{mikolajczyk2005performance, rana2016a}. To calculate the ROC curve, we need the values of precision (Equation~\ref{eq:precision}) and recall (Equation~\ref{eq:recall}), where tp, fp, and fn are, respectively, the number of true positives, false positives, and false negatives.

\begin{equation}
    P = \frac{\operatorname{tp}}{\operatorname{tp}+\operatorname{fp}}
    \label{eq:precision}
\end{equation}
\begin{equation}
    R = \frac{\operatorname{tp}}{\operatorname{tp}+\operatorname{fn}}
    \label{eq:recall}
\end{equation}

The ROC curve is obtained by calculating the precision and recall values for a sequence of \textit{th} values from the NNDR. The average APs for all permutations of image pairs in a dataset is called mean average precision (mAP).

\subsubsection{Algorithms and modifications}~\label{subsec:fp-algorithms}

\begin{figure*}[!htb]
    \centering
    \subfloat[\label{fig:modificacao_algoritmos_harris}]{%
        \includegraphics[width=0.90\linewidth]{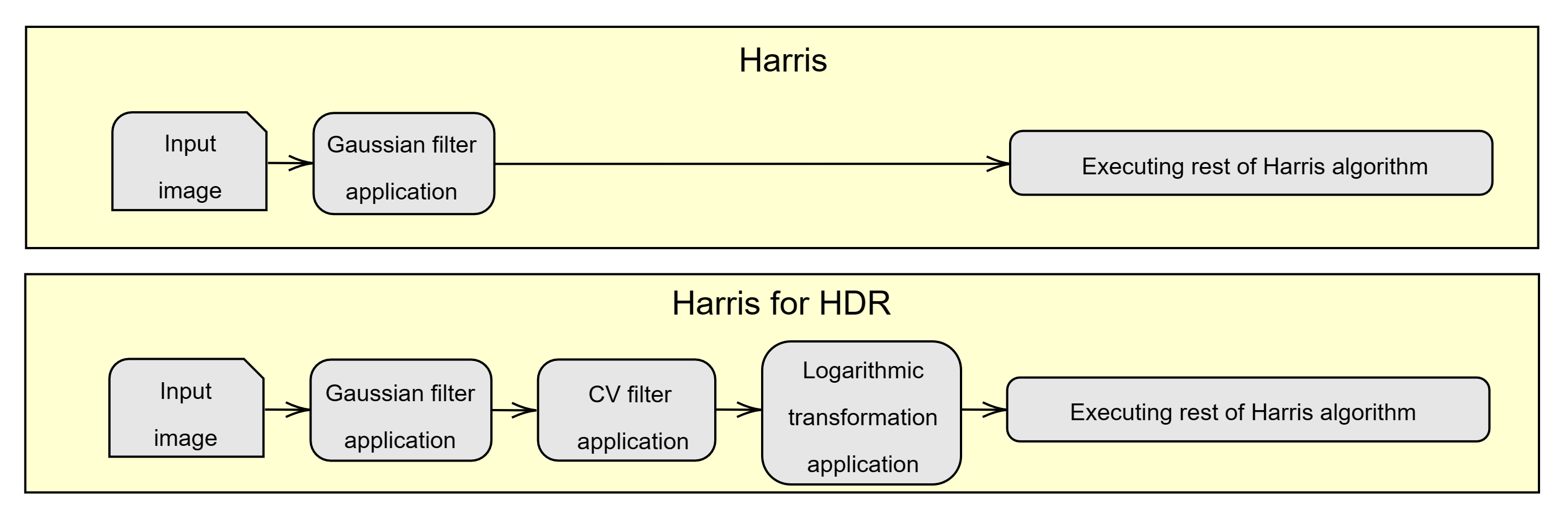}
    }
    \hfill
    \subfloat[\label{fig:modificacao_algoritmos_sift}]{%
        \includegraphics[width=0.90\linewidth]{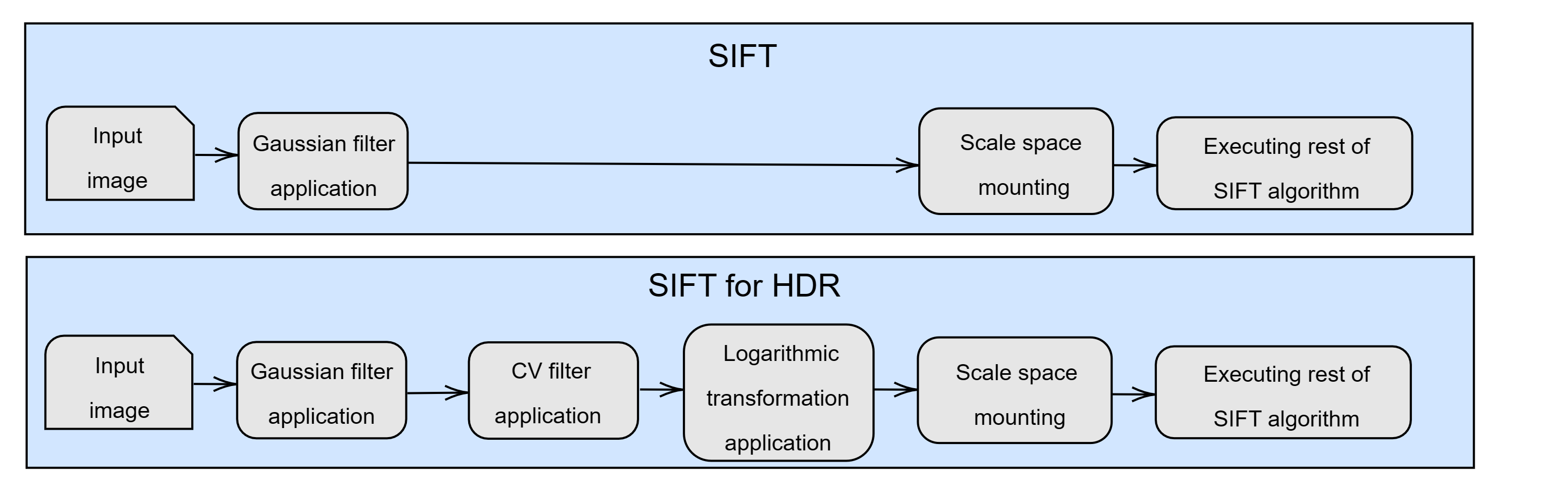}
    }
    \caption{Modification of the canonical detectors to implement the (a) Harris for HDR and (b) SIFT for HDR.}
    \label{fig:modificacao_algoritmos_harris_sift} 
\end{figure*}
\begin{figure}[t]
    \centering
    \subfloat[\label{fig:aplicacao_filtro_cv_original}]{%
        \includegraphics[width=0.3\linewidth]{l_3.jpg}
    }
    \hfill
    \subfloat[\label{fig:aplicacao_filtro_cv_original_ldr}]{%
        \includegraphics[width=0.3\linewidth]{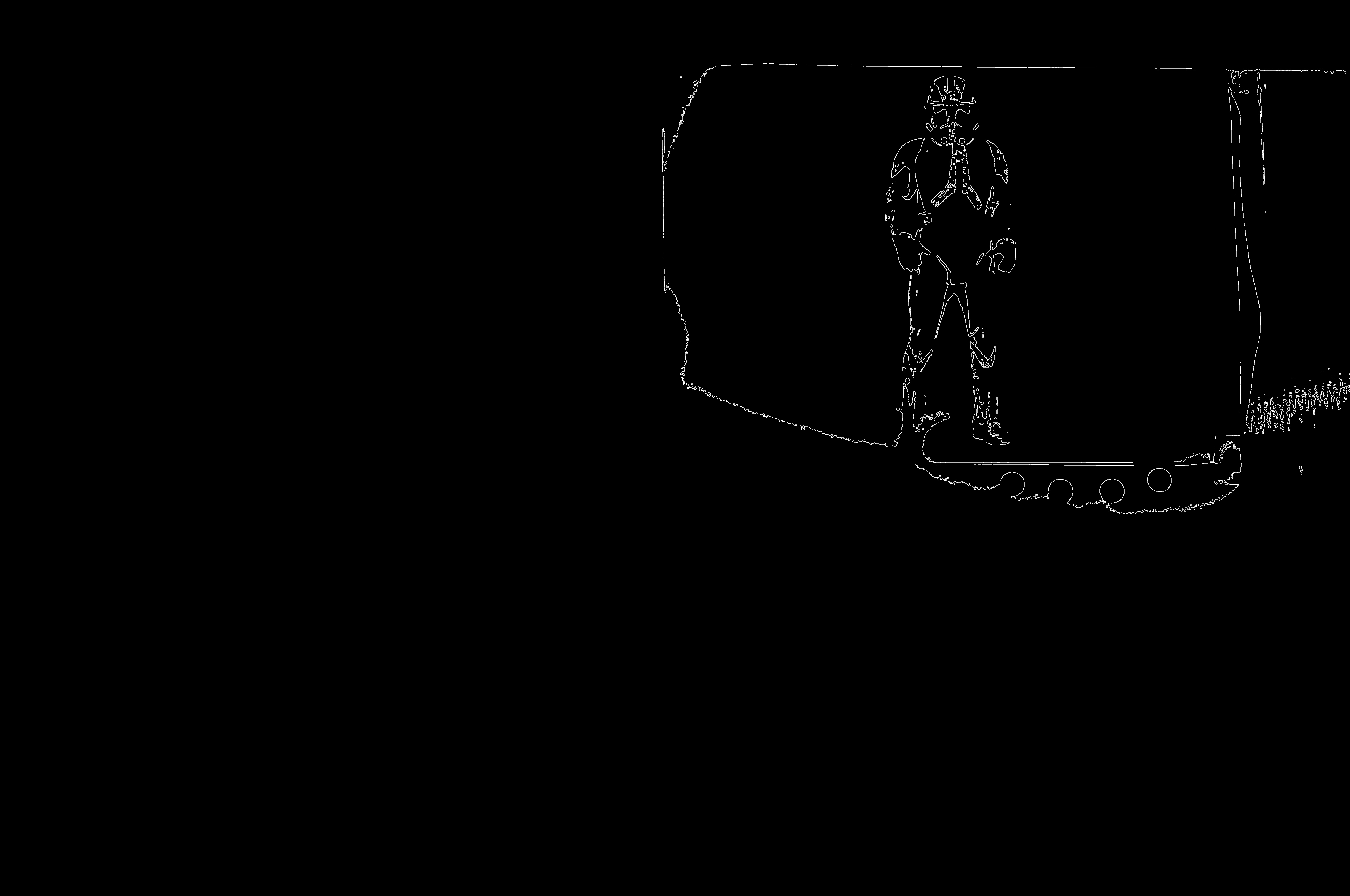}
    }
    \hfill
    \subfloat[\label{fig:aplicacao_filtro_cv_original_hdr}]{%
        \includegraphics[width=0.3\linewidth]{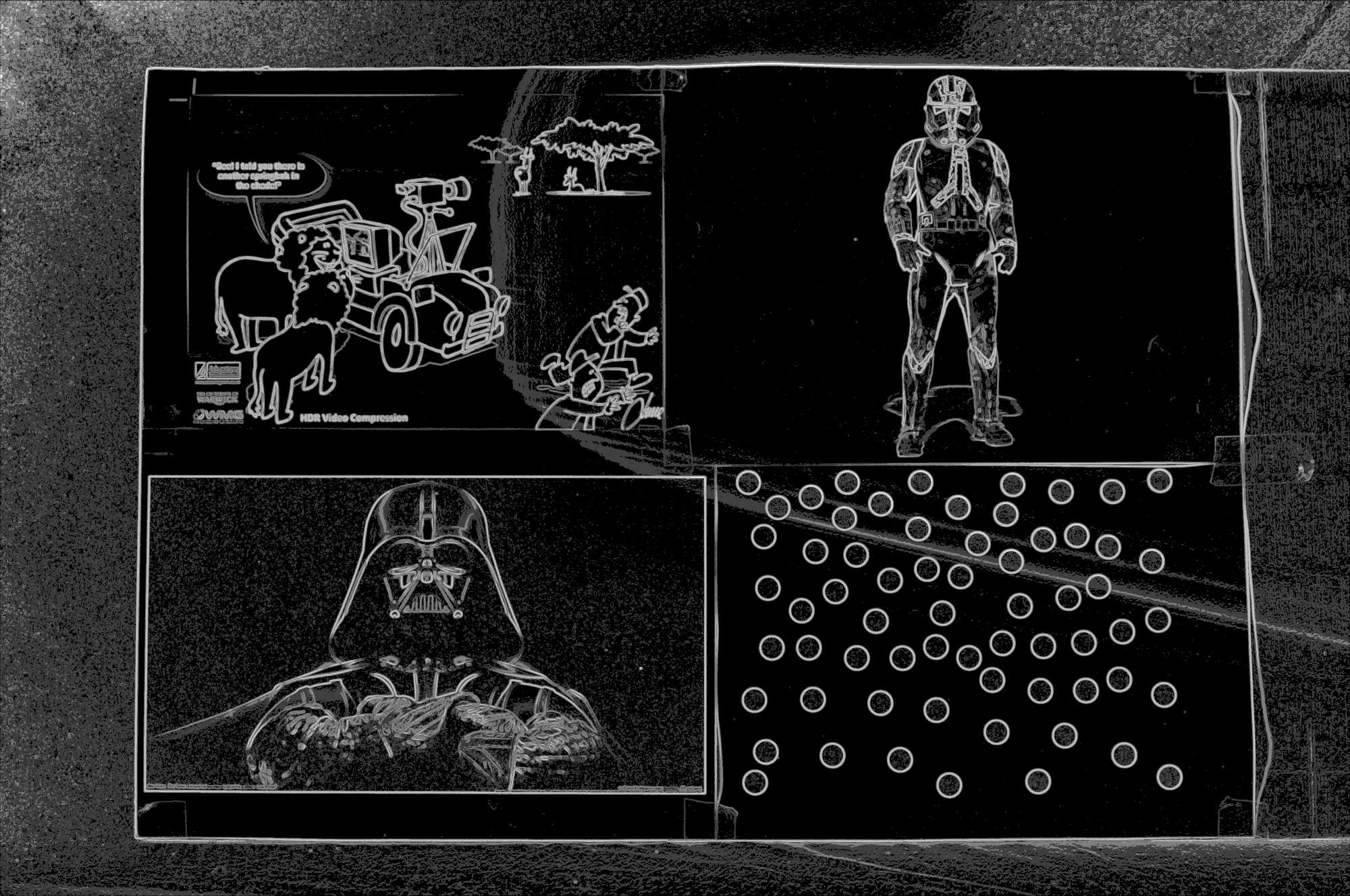}
    }
    \caption{Example of CV filter application on images: (a) reference image; (b) CV filter applied on a LDR capture; and (c) CV filter applied in a HDR capture.}
    \label{fig:exemplo_aplicacao_filtro_cv} 
\end{figure}
\begin{figure}[!htb]
    \centering
    \subfloat[\label{fig:dataset_rana_lr_segmentation_capture}]{%
        \includegraphics[width=0.235\linewidth]{lr_2.jpg}
    }
    \hfill
    \subfloat[\label{fig:dataset_rana_lr_segmentation_ROI}]{%
        \includegraphics[width=0.235\linewidth]{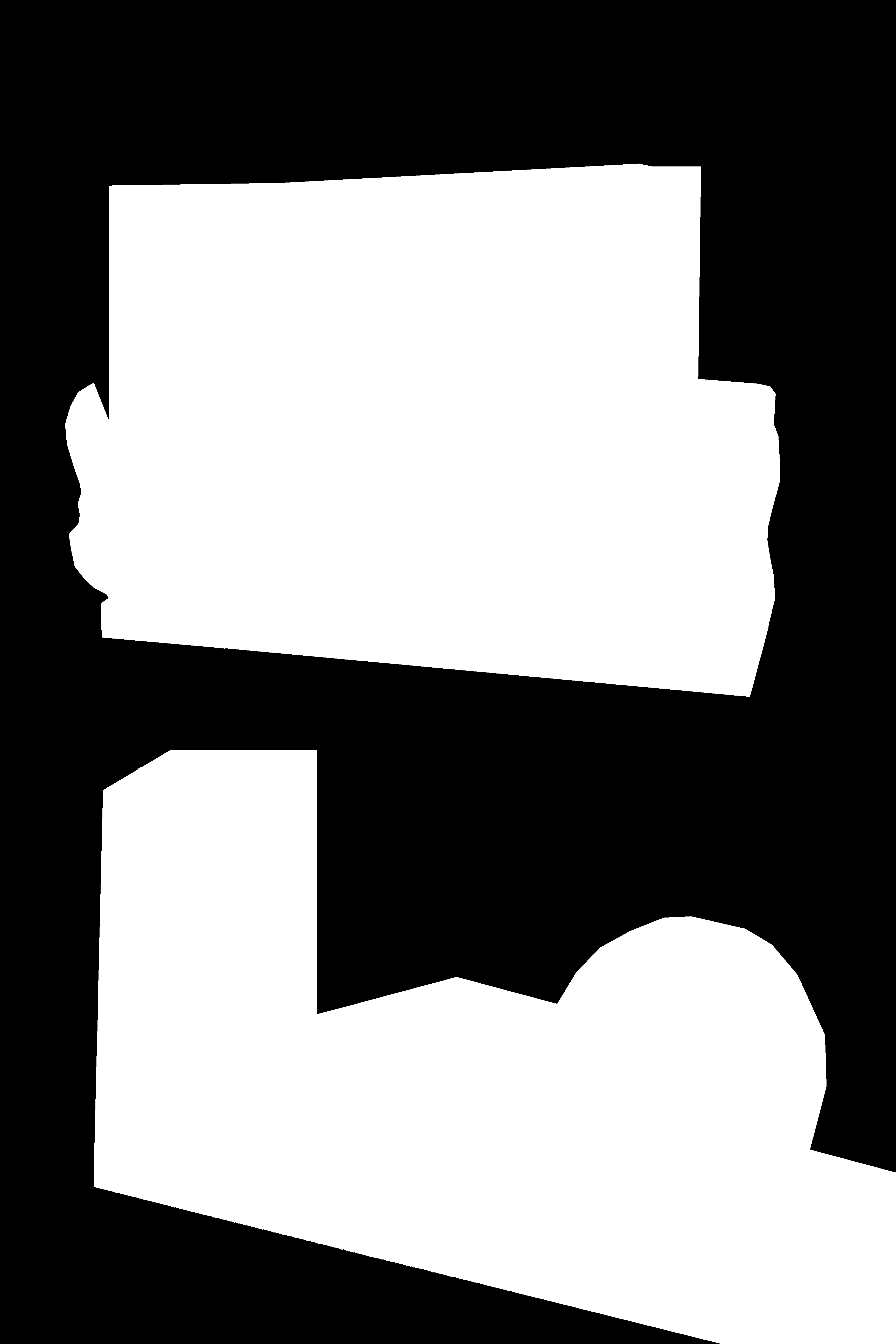}
    }
    \hfill
    \subfloat[\label{fig:dataset_rana_lr_segmentation_luminanceMap}]{%
        \includegraphics[width=0.235\linewidth]{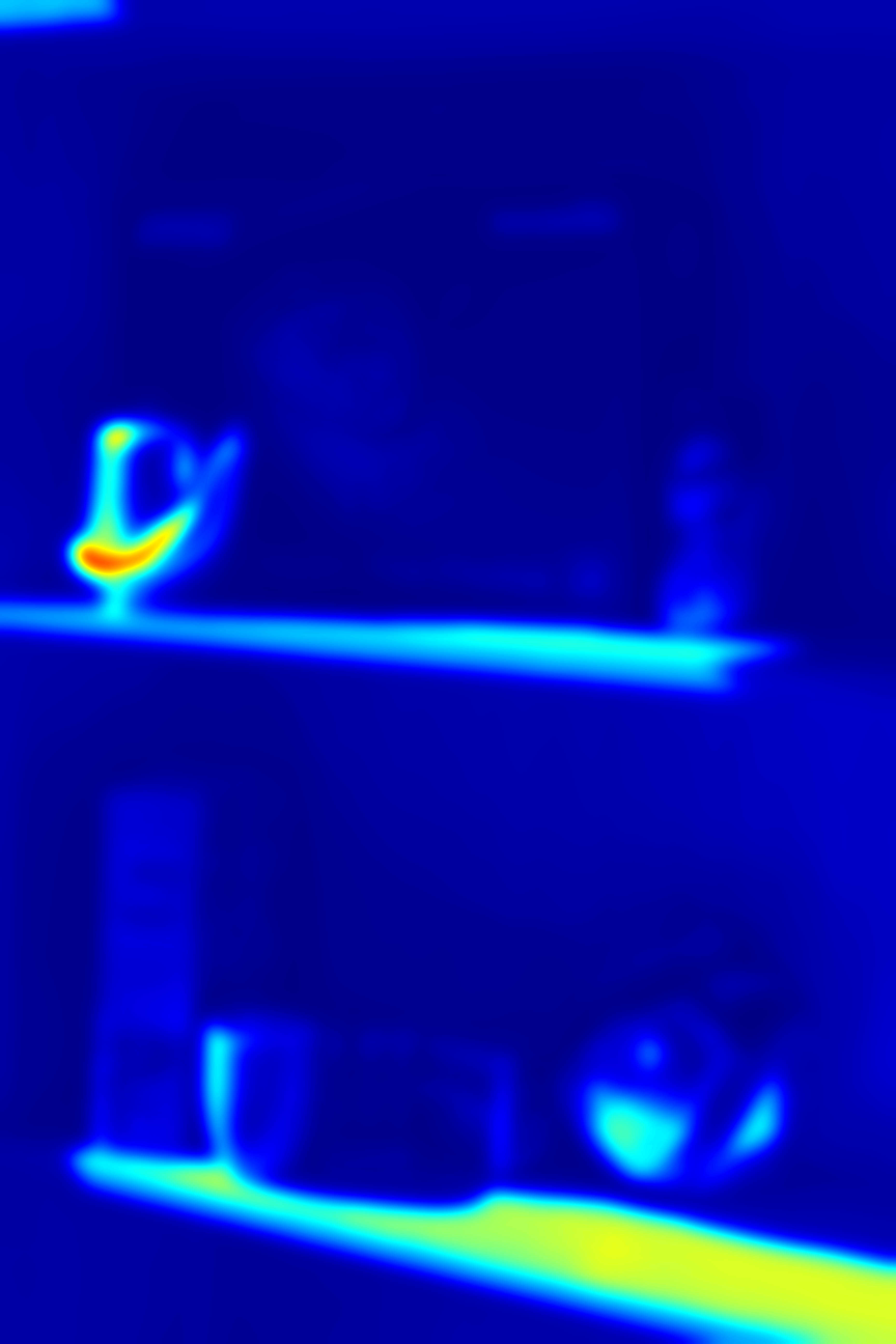}
    }
    \hfill
    \subfloat[\label{fig:dataset_rana_lr_segmentation_segments}]{%
        \includegraphics[width=0.235\linewidth]{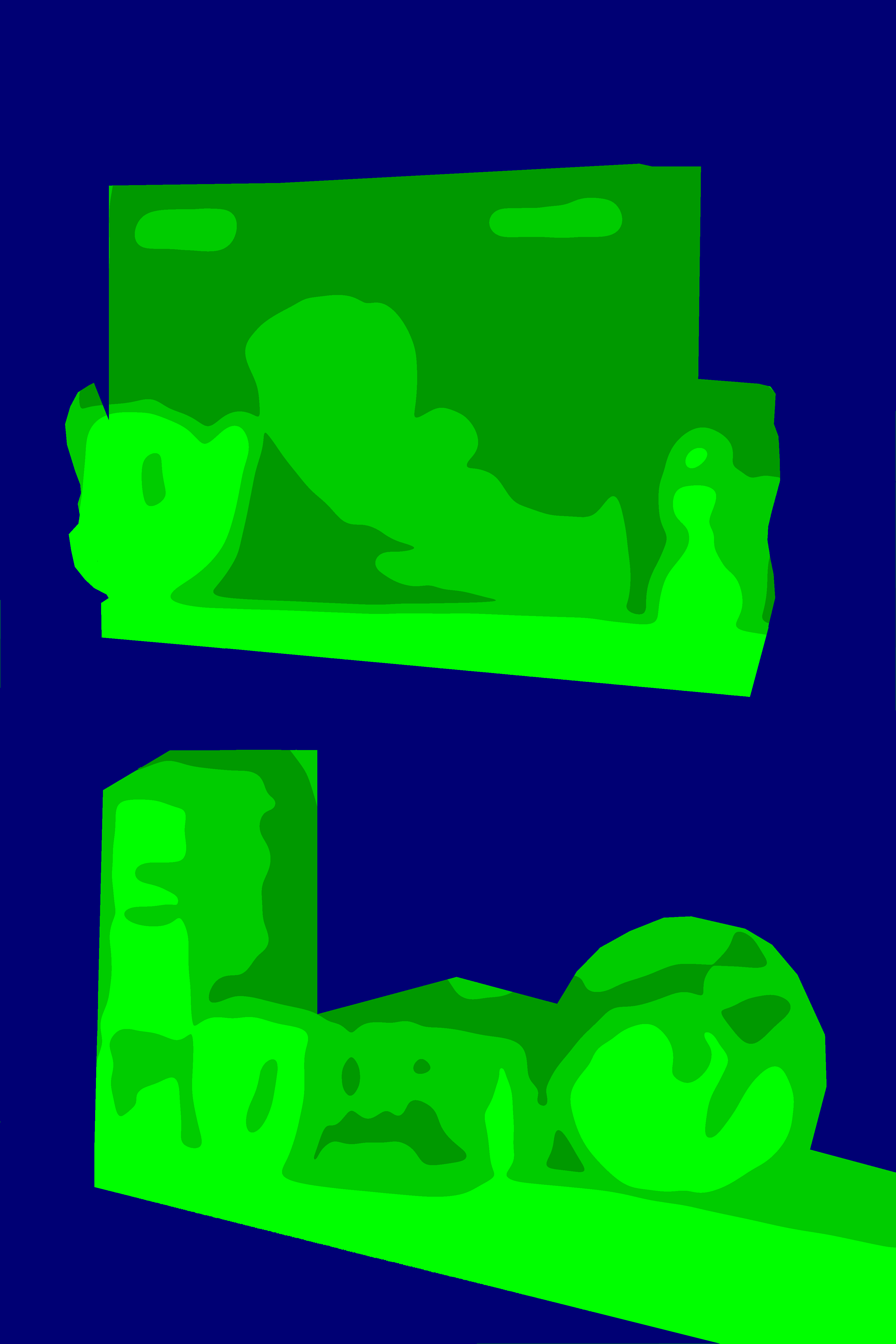}
    }
    \caption{Intensity segmentation example of DR dataset: (a) reference image; (b) the ROI mask that separates the background from the foreground; (c) the luminance map generated using the original image and rendered as a heat map; and (d) the final segmentation of the image's foreground.}
    \label{fig:segmentacao_rana_lr} 
\end{figure}
\begin{figure*}[!htb]
    \centering
    \includegraphics[width=0.90\textwidth]{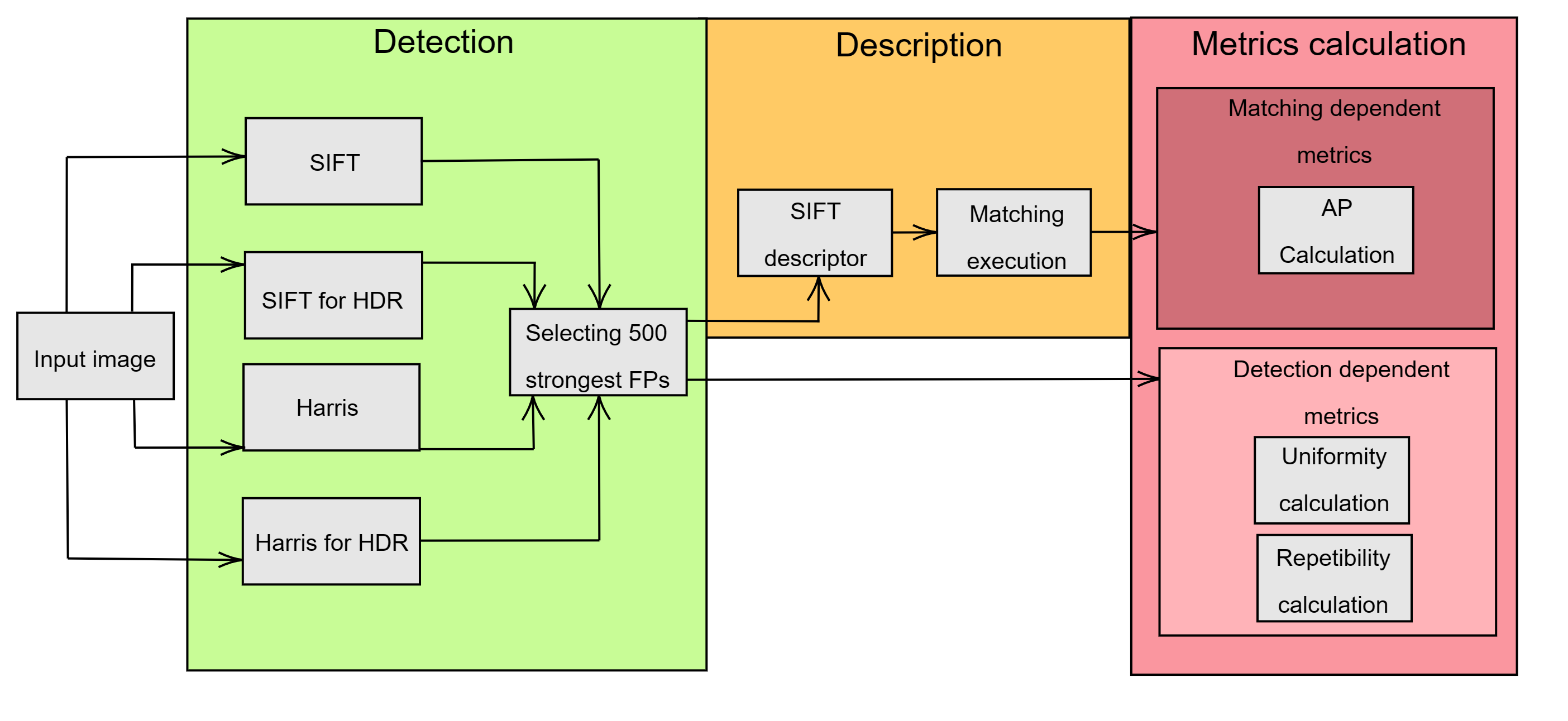}
    \caption{Execution pipeline. The green area refers to the FP detection, while the orange area refers to the description step. The red area refers to the metrics calculation step.}
    \label{fig:ilustracao_fluxo_execucao}
\end{figure*}

The modification in Harris corner and DoG proposed by Melo et al.~\cite{welerson_hdr} consists in adding a coefficient of variation (CV) filter in an intermediate step of the detection algorithm. The CV is defined as the ratio of the standard deviation ($\sigma$) to the arithmetic mean ($\mu$) of a given population, e.g., a set of pixels. Considering a population of $n$ elements, Equation~\ref{eq:cv} calculates the CV~\cite{livroEstatistica}. Figure~\ref{fig:exemplo_aplicacao_filtro_cv} shows an example of the CV filter application on LDR and HDR images.

\begin{equation}
    \operatorname{CV} = \frac{\sigma}{\mu} = \frac{\sqrt{\frac{1}{n}\sum_{i=1}^{n}(x_i-\mu)^2}}{\frac{1}{n}\sum_{i=1}^{n}x_i}
    \label{eq:cv}
\end{equation}

The coefficient of variation filter is used as an intermediate step in the HfHDR and SfHDR detectors to improve FP detection in darker image areas. In the HfHDR, the CV mask is applied to the input image after the Gaussian filter application step. After applying the CV filter, a logarithmic transformation is applied to the resulting image, and then algorithm continues as in the original Harris corner algorithm. In the SfHDR detector, a CV filter is applied to the input image followed by a logarithmic transformation before creating the scale space. Afterwards, the algorithm continues as in the original version by building the scale space. Figure~\ref{fig:modificacao_algoritmos_harris_sift} illustrates where Harris and SIFT algorithms were modified. For further details regarding the implementations of the CV filter and the logarithmic transformation, please refer to Melo et al. (2018)~\cite{welerson_hdr}.

\subsection{Datasets}~\label{subsec:ROIs}
        
Two datasets were created to explore extreme light conditions in scenes with LDR and HDR images. P{\v{r}}ibyl et al.~\cite{pvribyl2012} dataset (DP) was detailed in Section~\ref{subsubsec:tr-works_prybil} and Rana et al.~\cite{rana2015} dataset (DR) was detailed in Section~\ref{subsubsec:tr-works_rana}. Both datasets were used in several other studies~\cite{pvribyl2012, pvribyl2016, rana2015, rana2016a, rana2016b, rana2017a, rana2017b, rana2019, welerson_hdr, nascimento2022, nascimento2023}. In this study, we used the DR dataset and the three 2D capture sequences from DP dataset~\cite{pvribyl2012}.

\subsubsection{Dataset intensity segmentation}

The UR metric requires the segmentation of the image areas with different lighting intensities. The DP dataset provides ROIs to the image areas with high, medium, and low lighting, but the DP dataset does not provide it. We have implemented the following approach that segment an image in groups of lighting intensities, and can be used to segment any dataset scene. We used an approach similar to Nascimento et al. (2023)~\cite{nascimento2023}, described below.

First, we manually created a ROI to separate the background area from the foreground in the captures. The background is an area of the image with no features and where FPs are not usually detected. Each pixel in the foreground was assigned to one of three groups of lighting intensity: brightest, intermediate, and darkest pixels. We generated a luminance map for each capture to segment the image into the three groups. Then, we used the luminance map to select the brightest one-third of pixels, the intermediate one-third of pixels, and the darkest one-third of pixels. It resulted in three groups with a very similar amount of pixels, required for the UR metric. 
            
To generate the luminance map, a Retinex algorithm was used. According to the Retinex theory, an image $I$ is the product of the luminance $L$ of the scene and the reflectance $R$. That is, $I = RL$~\cite{rana2016b}. To find the luminance, we use the approach used by Chiu et al~\cite{tm_chiu1996}. The luminance is defined as $L = I \ast G_r$, and $G_r$ is a Gaussian filter with standard deviation calculated based on the image size $m \times n$, in the form: $r = \alpha \max(m,n)$. The $\alpha$ constant used was $0.007$, and the Gaussian mask size used was the smallest odd number greater than $6r$, as in previous studies~\cite{rana2016b, nascimento2022, nascimento2023}.

A cumulative histogram $H$ was calculated considering only the foreground in the luminance map. After that, we defined a threshold to divide the foreground into three groups, each one with approximately one-third of the pixels of the foreground. Figure~\ref{fig:dataset_rana_lr_segmentation_segments} shows an example of segmentation of the original image, with the background in blue, the darkest group of the image in dark green, the group with the intermediate incidence of illumination in medium green, and the brightest group of the image in light green.

Figure~\ref{fig:segmentacao_rana_lr} shows an example of the steps to segment an image. Figure~\ref{fig:dataset_rana_lr_segmentation_capture} shows the original image, while Figure~\ref{fig:dataset_rana_lr_segmentation_ROI} shows the ROI mask created manually to separate the foreground from the background. Figure~\ref{fig:dataset_rana_lr_segmentation_luminanceMap} shows the luminance map rendered as a heat map.

\subsection{Pipeline execution}~\label{subsec:pipeline}

To run the experiments, we used Rana et al. intensity segmented dataset. The P{\v{r}}ibyl et al. dataset already provides intensity segmentation. We used the FPs generated by Harris and Harris for HDR with SIFT descriptor to create the Harris+SIFT and HfHDR+SIFT descriptors.
        
We run the detection algorithms. Then, we classify all detected FPs based on their response to the detector. The 500 FPs with the strongest response were selected, as done in other studies~\cite{rana2015, welerson_hdr}. The selected FPs were described using SIFT descriptor. For each dataset, we matched the FPs all possible pairs of distinct images and calculated the metrics.

The UR, RR, AP, mAP, and matching rates were calculated for Harris, Harris for HDR, SIFT, SIFT for HDR, Harris+SIFT, and HfHDR+SIFT algorithms. Figure~\ref{fig:ilustracao_fluxo_execucao} illustrates the execution flow of the experiments. As mAP and matching rate metrics are calculated after the execution of the experiments, they are not present in the illustration.

\section{Results}~\label{sec:results}

Tables~\ref{tab:res-deteccao_srr_ur} to~\ref{tab:res-comparison_sift_melo} show the metric values obtained for each dataset, where 2D distance, 2D lighting, and 2D viewpoint are from P{\v{r}}ibyl et al.~\cite{pvribyl2012} dataset, and the ProjectRoom and LightRoom are from Rana et al. dataset~\cite{rana2015}.

\subsection{Detection evaluation}~\label{sec:results-detection}
        Table~\ref{tab:res-deteccao_srr_ur} shows the results obtained with sRR and UR mean for each dataset used in this study. The best result is highlighted.

\begin{table*}[!htb]
    \centering
    \caption{UR and sRR values using Harris, HfHDR, SIFT and SfHDR detectors for all datasets. The best results for each dataset are highlighted.}
    \label{tab:res-deteccao_srr_ur}
    \resizebox{0.8\textwidth}{!}
    {%
    \begin{tabular}{c|c|
            >{\columncolor[HTML]{EFEFEF}}l 
            >{\columncolor[HTML]{EFEFEF}}l |ll|
            >{\columncolor[HTML]{EFEFEF}}l 
            >{\columncolor[HTML]{EFEFEF}}l |ll|
        }
        \cline{2-10}
            & Detector & \multicolumn{2}{c|}{\cellcolor[HTML]{EFEFEF}Harris} & \multicolumn{2}{c|}{HfHDR} & \multicolumn{2}{c|}{\cellcolor[HTML]{EFEFEF}SIFT} & \multicolumn{2}{c|}{SfHDR} \\ 
        \cline{2-10} 
            & Database dynamic range & \multicolumn{1}{c|}{\cellcolor[HTML]{EFEFEF}LDR} & \multicolumn{1}{c|}{\cellcolor[HTML]{EFEFEF}HDR} & \multicolumn{1}{c|}{LDR} & \multicolumn{1}{c|}{HDR} & \multicolumn{1}{c|}{\cellcolor[HTML]{EFEFEF}LDR} & \multicolumn{1}{c|}{\cellcolor[HTML]{EFEFEF}HDR} & \multicolumn{1}{c|}{LDR} & \multicolumn{1}{c|}{HDR} \\ 
        \hline
            \multicolumn{1}{|c|}{\cellcolor[HTML]{EFEFEF}} & 2D distance & \multicolumn{1}{c|}{\cellcolor[HTML]{CBCEFA}\textbf{0,35581}} & \multicolumn{1}{c|}{\cellcolor[HTML]{EFEFEF}0,17330} & \multicolumn{1}{c|}{0,03914} & \multicolumn{1}{c|}{0,15857} & \multicolumn{1}{c|}{\cellcolor[HTML]{EFEFEF}0,20143} & \multicolumn{1}{c|}{\cellcolor[HTML]{EFEFEF}0,26066} & \multicolumn{1}{c|}{0,15981} & \multicolumn{1}{c|}{0,07161} \\
        \cline{2-10} 
            \multicolumn{1}{|c|}{\cellcolor[HTML]{EFEFEF}} & 2D lighting & \multicolumn{1}{c|}{\cellcolor[HTML]{EFEFEF}0,02041} & \multicolumn{1}{c|}{\cellcolor[HTML]{CBCEFA}\textbf{0,47553}} & \multicolumn{1}{c|}{0,00990} & \multicolumn{1}{c|}{0,01161} & \multicolumn{1}{c|}{\cellcolor[HTML]{EFEFEF}0,04101} & \multicolumn{1}{c|}{\cellcolor[HTML]{EFEFEF}0,01314} & \multicolumn{1}{c|}{0,01552} & \multicolumn{1}{c|}{0,00981} \\
        \cline{2-10}
            \multicolumn{1}{|c|}{\cellcolor[HTML]{EFEFEF}} & 2D viewpoint & \multicolumn{1}{c|}{\cellcolor[HTML]{EFEFEF}0,05168} & \multicolumn{1}{c|}{\cellcolor[HTML]{EFEFEF}0,05789} & \multicolumn{1}{c|}{0,02087} & \multicolumn{1}{c|}{0,02495} & \multicolumn{1}{c|}{\cellcolor[HTML]{CBCEFA}\textbf{0,14597}} & \multicolumn{1}{c|}{\cellcolor[HTML]{EFEFEF}0,04487} & \multicolumn{1}{c|}{0,03022} & \multicolumn{1}{c|}{0,01519} \\
        \cline{2-10} 
            \multicolumn{1}{|c|}{\cellcolor[HTML]{EFEFEF}} & LightRoom & \multicolumn{1}{c|}{\cellcolor[HTML]{CBCEFA}\textbf{0,41011}} & \multicolumn{1}{c|}{\cellcolor[HTML]{EFEFEF}0,32425} & \multicolumn{1}{c|}{0,05961} & 0,24409 & \multicolumn{1}{l|}{\cellcolor[HTML]{EFEFEF}0,16264} & 0,34942 & \multicolumn{1}{l|}{0,04981} & 0,22219 \\
        \cline{2-10} 
            \multicolumn{1}{|c|}{\multirow{-5}{*}{\cellcolor[HTML]{EFEFEF}
            \makecell{Mean\\repeatability\\(sRR)}}} & ProjectRoom & \multicolumn{1}{l|}{\cellcolor[HTML]{CBCEFA}\textbf{0,27704}} & 0,25020 & \multicolumn{1}{l|}{0,12238} & 0,07990 & \multicolumn{1}{l|}{\cellcolor[HTML]{EFEFEF}0,12734} & 0,13466 & \multicolumn{1}{l|}{0,02000} & 0,07209 \\
        \hline
        \hline
            \multicolumn{1}{|c|}{\cellcolor[HTML]{EFEFEF}} & 2D distance & \multicolumn{1}{l|}{\cellcolor[HTML]{EFEFEF}0,28204} & \cellcolor[HTML]{EFEFEF}0,35080 & \multicolumn{1}{l|}{0,08776} & 0,64780 & \multicolumn{1}{l|}{\cellcolor[HTML]{EFEFEF}0,24148} & 0,43342 & \multicolumn{1}{l|}{0,35257} & \cellcolor[HTML]{CBCEFA}\textbf{0,83200} \\
        \cline{2-10} 
            \multicolumn{1}{|c|}{\cellcolor[HTML]{EFEFEF}} & 2D lighting & \multicolumn{1}{l|}{\cellcolor[HTML]{EFEFEF}0,27465} & \cellcolor[HTML]{EFEFEF}0,15568 & \multicolumn{1}{l|}{0,24440} & 0,72771 & \multicolumn{1}{l|}{\cellcolor[HTML]{EFEFEF}0,16073} & 0,36771 & \multicolumn{1}{l|}{0,29542} & \cellcolor[HTML]{CBCEFA}\textbf{0,78000} \\
        \cline{2-10} 
            \multicolumn{1}{|c|}{\cellcolor[HTML]{EFEFEF}} & 2D viewpoint & \multicolumn{1}{l|}{\cellcolor[HTML]{EFEFEF}0,24951} & \cellcolor[HTML]{EFEFEF}0,23182 & \multicolumn{1}{l|}{0,05559} & 0,76610 & \multicolumn{1}{l|}{\cellcolor[HTML]{EFEFEF}0,14800} & 0,45657 & \multicolumn{1}{l|}{0,35285} & \cellcolor[HTML]{CBCEFA}\textbf{0,83285} \\
        \cline{2-10} 
            \multicolumn{1}{|c|}{\cellcolor[HTML]{EFEFEF}} & LightRoom & \multicolumn{1}{l|}{\cellcolor[HTML]{EFEFEF}0,40260} & \cellcolor[HTML]{EFEFEF}0,10505 & \multicolumn{1}{l|}{0,54991} & \cellcolor[HTML]{CBCEFA}\textbf{0,83878} & \multicolumn{1}{l|}{\cellcolor[HTML]{EFEFEF}0,20640} & 0,19238 & \multicolumn{1}{l|}{0,21314} & 0,68348 \\
        \cline{2-10} 
            \multicolumn{1}{|c|}{\multirow{-5}{*}{\cellcolor[HTML]{EFEFEF}\makecell{Mean\\uniformity\\(UR)}}} & ProjectRoom & \multicolumn{1}{l|}{\cellcolor[HTML]{EFEFEF}0,58157} & \cellcolor[HTML]{EFEFEF}0,17471 & \multicolumn{1}{l|}{\cellcolor[HTML]{CBCEFA}\textbf{0,81046}} & 0,77689 & \multicolumn{1}{l|}{\cellcolor[HTML]{EFEFEF}0,68581} & 0,37095 & \multicolumn{1}{l|}{0,65569} & 0,72779 \\
        \hline
    \end{tabular}
    }%
\end{table*}
        
\begin{figure}[!htb]
    \centering
    \subfloat[\label{fig:compara_deteccao_ldr_harrisForHDR}]{%
        \includegraphics[width=0.48\columnwidth]{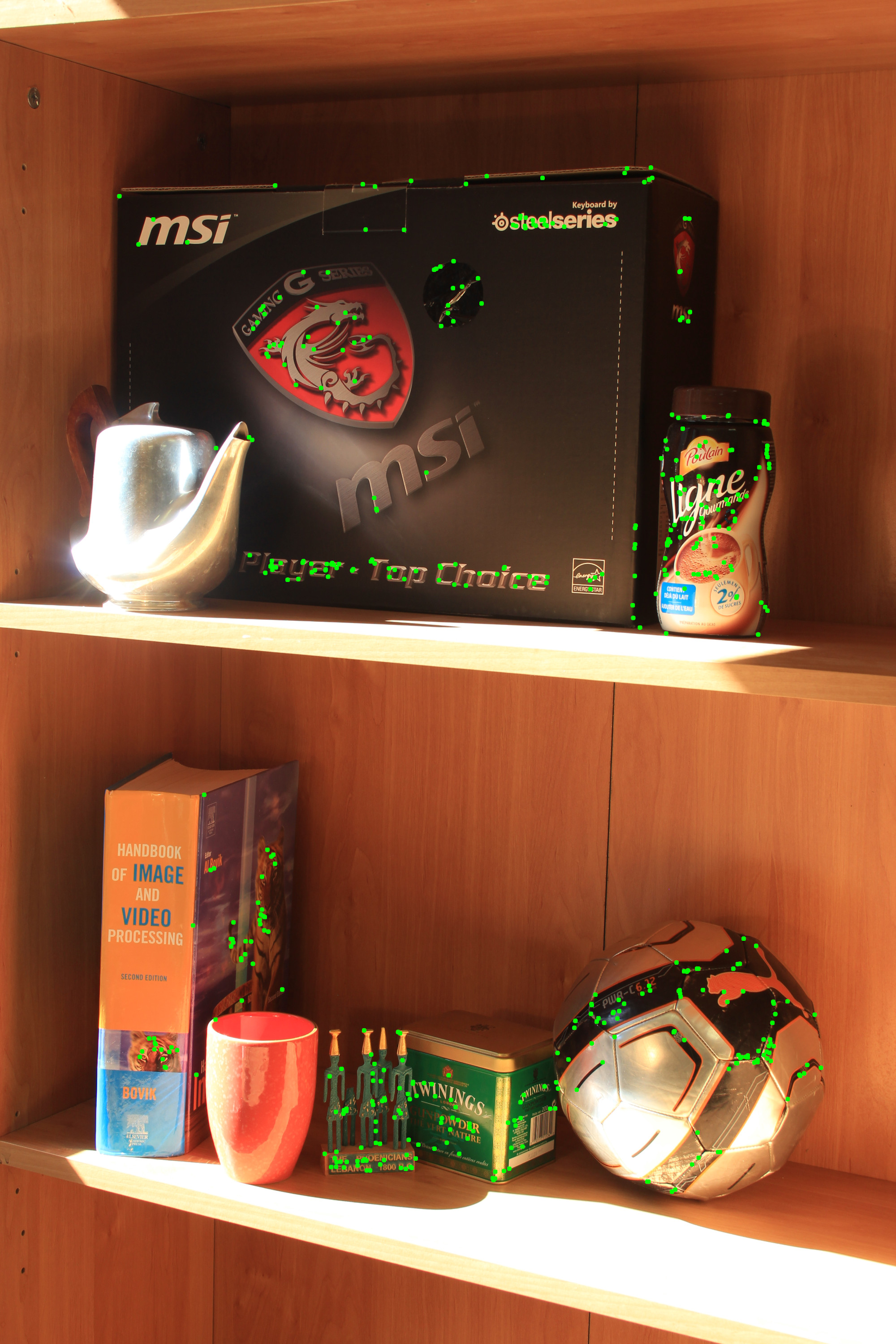}
    }
    \hfill
    \subfloat[\label{fig:compara_deteccao_hdr_harrisForHDR}]{%
        \includegraphics[width=0.48\columnwidth]{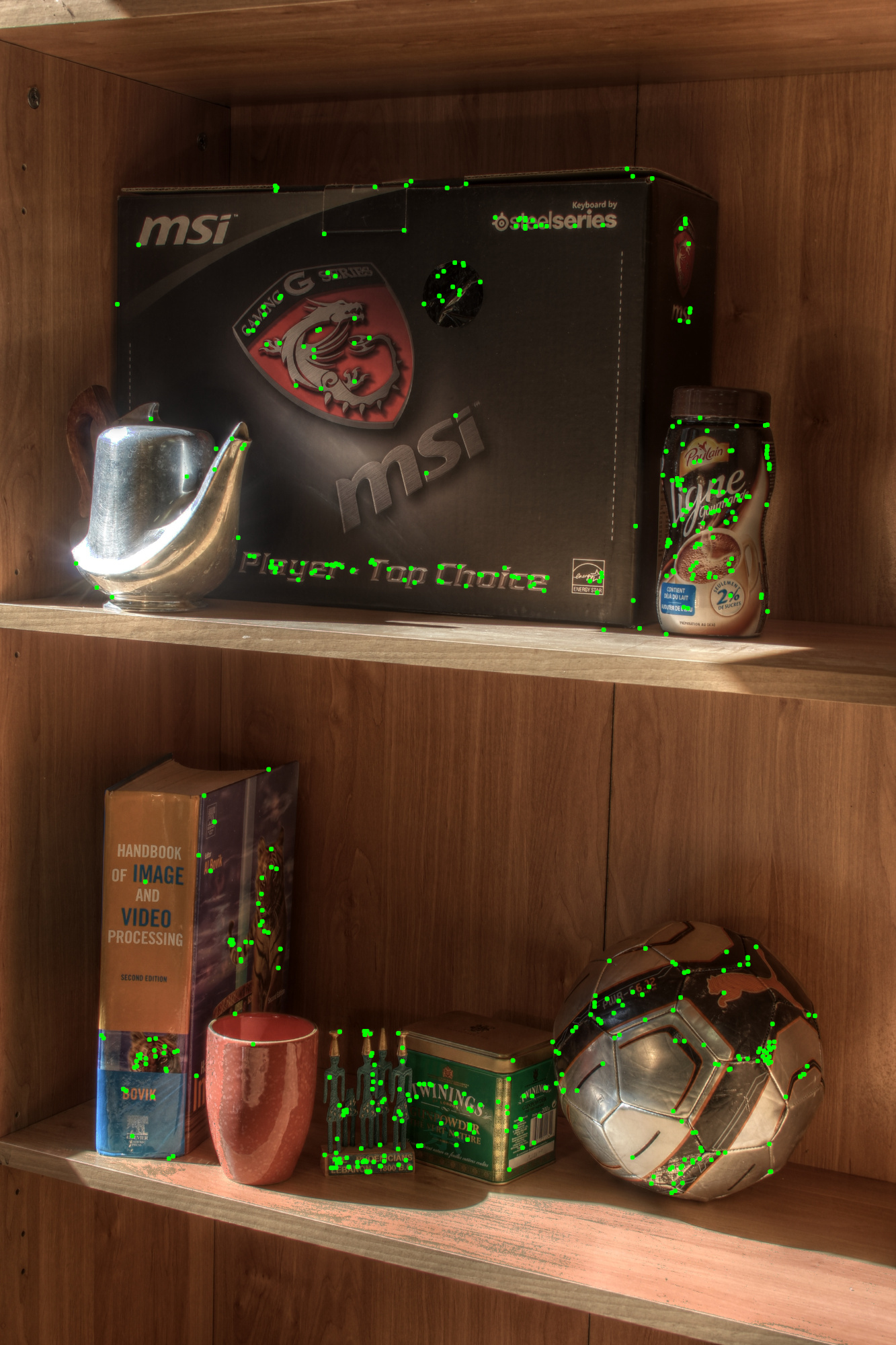}
    }
    \caption{FP detection using LightRoom dataset and HfHDR detector. (a) using LDR image and (b) using HDR image. For display purposes, HDR image was tone-mapped to LDR.}
    \label{fig:res-exemplo_deteccao_rana} 
\end{figure}

The Harris and SIFT algorithms presented better sRR values, but the HfHDR and SfHDR presented better UR values. SfHDR showed better UR values in three out of the five datasets. In most cases, the UR metric has better values when using HDR images. Figure~\ref{fig:res-exemplo_deteccao_rana} shows an example of FP detection using the LightRoom dataset and HfHDR algorithm. There, we can observe the improvement in the FP distribution when using HDR images, especially in darker and intermediate areas.

\subsection{Description evaluation}~\label{sec:resultados-descricao}

Table~\ref{tab:res-descricao_map_matching} shows the mAP and the mean matching rate for each dataset used in this study. The best result is highlighted.

\begin{table*}[!htb]
    \centering
    \caption{Mean average precision (mAP) and matching rate values using Harris+SIFT, HfHDR+SIFT, SIFT and SfHDR descriptors for all datasets. The best result for each dataset is highlighted.}
    \label{tab:res-descricao_map_matching}
    \resizebox{0.8\textwidth}{!}
    {%
    \begin{tabular}{c|c|
            >{\columncolor[HTML]{EFEFEF}}l 
            >{\columncolor[HTML]{EFEFEF}}l |ll|
            >{\columncolor[HTML]{EFEFEF}}l 
            >{\columncolor[HTML]{EFEFEF}}l |ll|
        }
        \cline{2-10}
            & Descriptor & \multicolumn{2}{c|}{\cellcolor[HTML]{EFEFEF}Harris+SIFT} & \multicolumn{2}{c|}{HfHDR+SIFT} & \multicolumn{2}{c|}{\cellcolor[HTML]{EFEFEF}SIFT} & \multicolumn{2}{c|}{SfHDR} \\
        \cline{2-10} 
            & Database dynamic range & \multicolumn{1}{c|}{\cellcolor[HTML]{EFEFEF}LDR} & \multicolumn{1}{c|}{\cellcolor[HTML]{EFEFEF}HDR} & \multicolumn{1}{c|}{LDR} & \multicolumn{1}{c|}{HDR} & \multicolumn{1}{c|}{\cellcolor[HTML]{EFEFEF}LDR} & \multicolumn{1}{c|}{\cellcolor[HTML]{EFEFEF}HDR} & \multicolumn{1}{c|}{LDR} & \multicolumn{1}{c|}{HDR} \\
        \hline
            \multicolumn{1}{|c|}{\cellcolor[HTML]{EFEFEF}} & 2D distance & \multicolumn{1}{l|}{\cellcolor[HTML]{EFEFEF}0,68872} & 0,39798 & \multicolumn{1}{l|}{0,29657} & 0,75337 & \multicolumn{1}{l|}{\cellcolor[HTML]{EFEFEF}0,44677} & \cellcolor[HTML]{CBCEFA}\textbf{0,82056} & \multicolumn{1}{l|}{0,45285} & 0,51048 \\
        \cline{2-10} 
            \multicolumn{1}{|c|}{\cellcolor[HTML]{EFEFEF}} & 2D lighting & \multicolumn{1}{l|}{\cellcolor[HTML]{EFEFEF}0,32830} & \cellcolor[HTML]{CBCEFA}\textbf{0,35886} & \multicolumn{1}{l|}{0,13837} & 0,31867 & \multicolumn{1}{l|}{\cellcolor[HTML]{EFEFEF}0,21669} & \cellcolor[HTML]{EFEFEF}0,33776 & \multicolumn{1}{l|}{0,27214} & 0,23069 \\
        \cline{2-10}
            \multicolumn{1}{|c|}{\cellcolor[HTML]{EFEFEF}} & 2D viewpoint & \multicolumn{1}{l|}{\cellcolor[HTML]{EFEFEF}0,45687} & 0,16820 & \multicolumn{1}{l|}{0,00515} & \cellcolor[HTML]{CBCEFA}\textbf{0,51096} & \multicolumn{1}{l|}{\cellcolor[HTML]{EFEFEF}0,46316} & 0,46357 & \multicolumn{1}{l|}{0,43244} & 0,38100 \\
        \cline{2-10} 
            \multicolumn{1}{|c|}{\cellcolor[HTML]{EFEFEF}} & LightRoom & \multicolumn{1}{l|}{\cellcolor[HTML]{EFEFEF}0,50288} & 0,35428 & \multicolumn{1}{l|}{0,23895} & \cellcolor[HTML]{CBCEFA}\textbf{0,55231} & \multicolumn{1}{l|}{\cellcolor[HTML]{EFEFEF}0,16552} & 0,49372 & \multicolumn{1}{l|}{0,11931} & 0,46522 \\
        \cline{2-10} 
            \multicolumn{1}{|c|}{\multirow{-5}{*}{\cellcolor[HTML]{EFEFEF}\makecell{Mean\\average\\precision\\(mAP)}}} & ProjectRoom & \multicolumn{1}{l|}{\cellcolor[HTML]{CBCEFA}\textbf{0,66438}} & 0,32320 & \multicolumn{1}{l|}{0,27934} & 0,37020 & \multicolumn{1}{l|}{\cellcolor[HTML]{EFEFEF}0,30082} & 0,28835 & \multicolumn{1}{l|}{0,11896} & 0,28746 \\
        \hline
        \hline
            \multicolumn{1}{|c|}{\cellcolor[HTML]{EFEFEF}} & 2D distance & \multicolumn{1}{l|}{\cellcolor[HTML]{EFEFEF}0,96028} & \cellcolor[HTML]{EFEFEF}0,53647 & \multicolumn{1}{l|}{0,42108} & 0,91779 & \multicolumn{1}{l|}{\cellcolor[HTML]{EFEFEF}0,67077} & \cellcolor[HTML]{CBCEFA}\textbf{0,98200} & \multicolumn{1}{l|}{0,66566} & 0,69631 \\
        \cline{2-10} 
            \multicolumn{1}{|c|}{\cellcolor[HTML]{EFEFEF}} & 2D lighting & \multicolumn{1}{l|}{\cellcolor[HTML]{EFEFEF}0,47318} & \cellcolor[HTML]{CBCEFA}\textbf{0,50426} & \multicolumn{1}{l|}{0,20420} & 0,44893 & \multicolumn{1}{l|}{\cellcolor[HTML]{EFEFEF}0,32642} & 0,49055 & \multicolumn{1}{l|}{0,40935} & 0,36054 \\
        \cline{2-10} 
            \multicolumn{1}{|c|}{\cellcolor[HTML]{EFEFEF}} & 2D viewpoint & \multicolumn{1}{l|}{\cellcolor[HTML]{EFEFEF}0,73680} & \cellcolor[HTML]{EFEFEF}0,27323 & \multicolumn{1}{l|}{0,00896} & \cellcolor[HTML]{CBCEFA}\textbf{0,82050} & \multicolumn{1}{l|}{\cellcolor[HTML]{EFEFEF}0,78493} & 0,74640 & \multicolumn{1}{l|}{0,71270} & 0,63074 \\
        \cline{2-10} 
            \multicolumn{1}{|c|}{\cellcolor[HTML]{EFEFEF}} & LightRoom & \multicolumn{1}{l|}{\cellcolor[HTML]{EFEFEF}0,63017} & \cellcolor[HTML]{EFEFEF}0,48615 & \multicolumn{1}{l|}{0,31616} & \cellcolor[HTML]{CBCEFA}\textbf{0,70434} & \multicolumn{1}{l|}{\cellcolor[HTML]{EFEFEF}0,22440} & 0,69950 & \multicolumn{1}{l|}{0,16511} & 0,65916 \\
        \cline{2-10} 
            \multicolumn{1}{|c|}{\multirow{-5}{*}{\cellcolor[HTML]{EFEFEF}\makecell{Matching\\rate}}} & ProjectRoom & \multicolumn{1}{l|}{\cellcolor[HTML]{CBCEFA}\textbf{0,76277}} & \cellcolor[HTML]{EFEFEF}0,37288 & \multicolumn{1}{l|}{0,33243} & 0,40658 & \multicolumn{1}{l|}{\cellcolor[HTML]{EFEFEF}0,40326} & 0,36927 & \multicolumn{1}{l|}{0,16329} & 0,39886 \\
        \hline
    \end{tabular}
    }%
\end{table*}
    
Both mAP and matching rate presented similar results in each database. The Harris+SIFT and HfHDR+SIFT algorithms showed better results than SIFT and SfHDR. The SIFT algorithm performs better when using 2D distance dataset, probably due to SIFT's scale-invariance. In almost all cases, the best results were obtained using HDR images. Overall, the HfHDR+SIFT algorithm showed improvement in the description when compared to Harris+SIFT, and the SfHDR showed a worse description when compared to SIFT.
        
While P{\v{r}}ibyl et al. datasets contain images of a poster, Rana et al. datasets consist of scenes composed of several rigid objects positioned in such a way as to create shadows on other objects. In all metrics, when using the Rana datasets, the best results were obtained with the Harris or HfHDR algorithms. This may indicate that, in scenes with more complex lighting or geometry, corner detectors have an advantage over blob detectors.
        
The use of the CV filter and the logarithmic transformation in HDR images generate noise in the darkest regions and on the specular surfaces. But, when the CV filter is applied to LDR images, the resulting image showed edges only in the brightest regions of the image. Considering that the ProjectRoom dataset presented noisiest foreground when using HDR images and SfHDR and HfHDR+SIFT descriptors, we assume that this is why the best results in the ProjectRoom are obtained when using LDR images.
        
\begin{figure}[!htb]
    \centering
    \subfloat[\label{fig:matching_harrisForHdr_LDR}]{%
        \includegraphics[width=0.98\columnwidth]{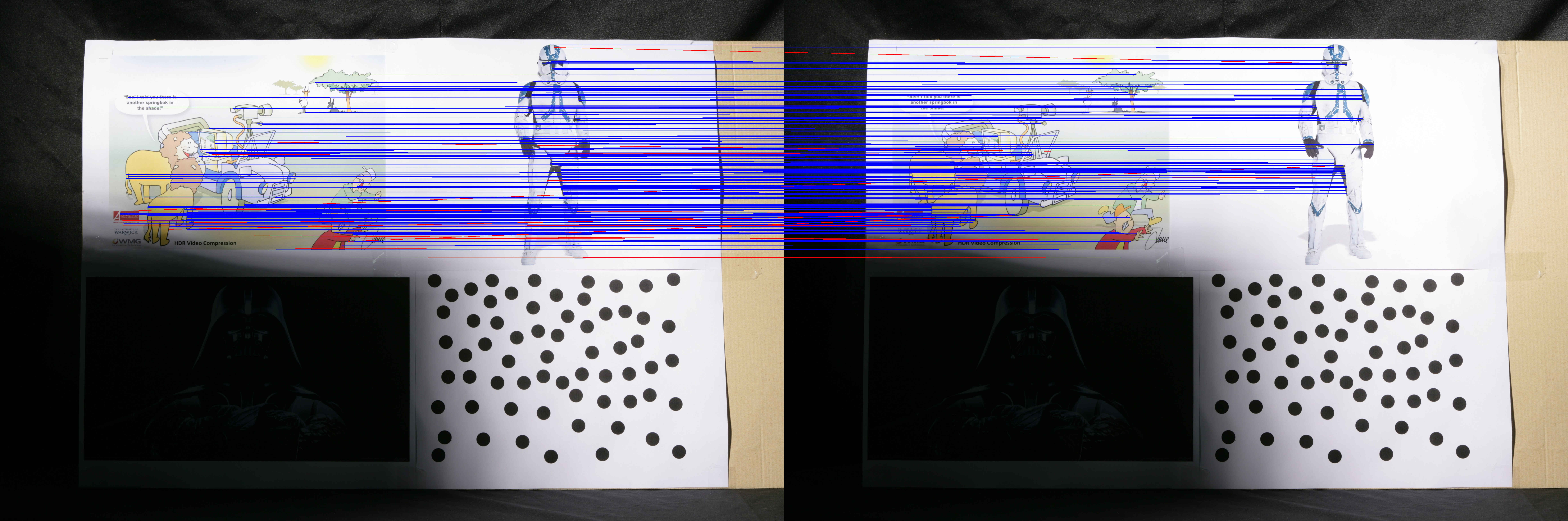}
    }
    \hfill
    \subfloat[\label{fig:matching_siftForHdr_HDR}]{%
        \includegraphics[width=0.98\columnwidth]{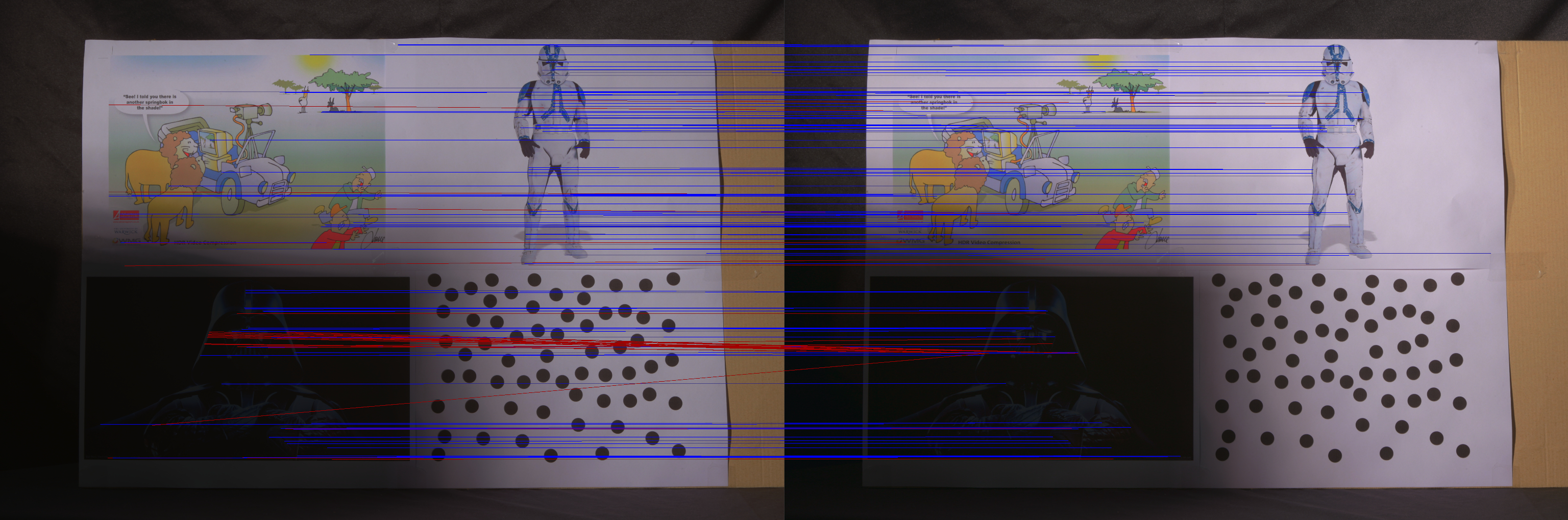}
    }
    \caption{Matching using SfHDR descriptions. Blue lines are correct matches, while red lines are incorrect matches. (a) using LDR image and (b) using HDR image.}
    \label{fig:res-exemplo_deteccao_pribyl_matching}
\end{figure}

Figure~\ref{fig:res-exemplo_deteccao_pribyl_matching} shows the FP matching using SfHDR algorithm. When HDR images are used, there are FP matches in the darkest area, while when using LDR images, there are no matches. Consequently, matching distribution between areas is improved when using HDR images. The UR metric confirms this, since most of the best URs were obtained when using SfHDR and HfHDR algorithms with HDR images.


\subsection{Comparison with previous studies}~\label{sec:res-comparison_previous_works}
        
Tables~\ref{tab:res-comparison_harris_melo} and~\ref{tab:res-comparison_sift_melo} compare Melo et al.~\cite{welerson_hdr} results with ours, using sRR and UR metrics in the 2D distance, 2D lighting, and 2D viewpoint datasets. It is important to note that SIFT and SfHDR detector algorithms were significantly improved, with subpixel precision and code rewriting to fix several problems of code cohesion and minor bug fixes. This can explain the difference between results.

\begin{table*}[!htb]
    \centering
    \caption{Comparison between Melo et al. and this study using Harris and HfHDR detectors. The best result for each dataset is highlighted.}
    \label{tab:res-comparison_harris_melo}
    \resizebox{0.8\textwidth}{!}
    {%
    \begin{tabular}{cc|
            >{\columncolor[HTML]{EFEFEF}}l 
            >{\columncolor[HTML]{EFEFEF}}l 
            >{\columncolor[HTML]{EFEFEF}}l 
            >{\columncolor[HTML]{EFEFEF}}l |
            >{\columncolor[HTML]{FFFFFF}}l 
            >{\columncolor[HTML]{FFFFFF}}l 
            >{\columncolor[HTML]{FFFFFF}}l 
            >{\columncolor[HTML]{FFFFFF}}l |}
        \cline{2-10}
            \multicolumn{1}{c|}{} & Detector & \multicolumn{2}{c|}{\cellcolor[HTML]{EFEFEF}Harris} & \multicolumn{2}{c|}{\cellcolor[HTML]{EFEFEF}HfHDR} & \multicolumn{2}{c|}{\cellcolor[HTML]{FFFFFF}Harris} & \multicolumn{2}{c|}{\cellcolor[HTML]{FFFFFF}HfHDR} \\
        \cline{2-10} 
            \multicolumn{1}{c|}{} & Database dynamic range & \multicolumn{1}{c|}{\cellcolor[HTML]{EFEFEF}LDR} & \multicolumn{1}{c|}{\cellcolor[HTML]{EFEFEF}HDR} & \multicolumn{1}{c|}{\cellcolor[HTML]{EFEFEF}LDR} & \multicolumn{1}{c|}{\cellcolor[HTML]{EFEFEF}HDR} & \multicolumn{1}{c|}{\cellcolor[HTML]{FFFFFF}LDR} & \multicolumn{1}{c|}{\cellcolor[HTML]{FFFFFF}HDR} & \multicolumn{1}{c|}{\cellcolor[HTML]{FFFFFF}LDR} & \multicolumn{1}{c|}{\cellcolor[HTML]{FFFFFF}HDR} \\
        \hline
            \multicolumn{1}{|c|}{\cellcolor[HTML]{EFEFEF}} & 2D distance & \multicolumn{1}{l|}{\cellcolor[HTML]{CBCEFA}\textbf{0,64000}} & \multicolumn{1}{l|}{\cellcolor[HTML]{EFEFEF}0,00000} & \multicolumn{1}{l|}{\cellcolor[HTML]{EFEFEF}0,00000} & 0,37000 & \multicolumn{1}{l|}{\cellcolor[HTML]{FFFFFF}0,01835} & \multicolumn{1}{l|}{\cellcolor[HTML]{FFFFFF}0,45263} & \multicolumn{1}{l|}{\cellcolor[HTML]{FFFFFF}0,00866} & 0,01009 \\
        \cline{2-10} 
            \multicolumn{1}{|c|}{\cellcolor[HTML]{EFEFEF}} & 2D lighting & \multicolumn{1}{l|}{\cellcolor[HTML]{CBCEFA}\textbf{0,70000}} & \multicolumn{1}{l|}{\cellcolor[HTML]{EFEFEF}0,00000} & \multicolumn{1}{l|}{\cellcolor[HTML]{EFEFEF}0,00000} & 0,43000 & \multicolumn{1}{l|}{\cellcolor[HTML]{FFFFFF}0,33066} & \multicolumn{1}{l|}{\cellcolor[HTML]{FFFFFF}0,17308} & \multicolumn{1}{l|}{\cellcolor[HTML]{FFFFFF}0,03638} & 0,14304 \\
        \cline{2-10} 
            \multicolumn{1}{|c|}{\multirow{-3}{*}{\cellcolor[HTML]{EFEFEF}\makecell{Mean\\repeatability\\(sRR)}}} & 2D viewpoint & \multicolumn{1}{l|}{\cellcolor[HTML]{EFEFEF}0,00000} & \multicolumn{1}{l|}{\cellcolor[HTML]{EFEFEF}0,00000} & \multicolumn{1}{l|}{\cellcolor[HTML]{EFEFEF}0,00000} & \cellcolor[HTML]{CBCEFA}\textbf{0,49000} & \multicolumn{1}{l|}{\cellcolor[HTML]{FFFFFF}0,04614} & \multicolumn{1}{l|}{\cellcolor[HTML]{FFFFFF}0,05324} & \multicolumn{1}{l|}{\cellcolor[HTML]{FFFFFF}0,01822} & 0,02180 \\
        \hline
        \hline
            \multicolumn{1}{|c|}{\cellcolor[HTML]{EFEFEF}} & 2D distance & \multicolumn{1}{l|}{\cellcolor[HTML]{EFEFEF}0,27000} & \multicolumn{1}{l|}{\cellcolor[HTML]{EFEFEF}0,39000} & \multicolumn{1}{l|}{\cellcolor[HTML]{EFEFEF}0,06000} & 0,42000 & \multicolumn{1}{l|}{\cellcolor[HTML]{FFFFFF}0,28204} & \multicolumn{1}{l|}{\cellcolor[HTML]{FFFFFF}0,35080} & \multicolumn{1}{l|}{\cellcolor[HTML]{FFFFFF}0,08776} & \cellcolor[HTML]{CBCEFA}\textbf{0,64780} \\
        \cline{2-10} 
            \multicolumn{1}{|c|}{\cellcolor[HTML]{EFEFEF}} & 2D lighting & \multicolumn{1}{l|}{\cellcolor[HTML]{EFEFEF}0,35000} & \multicolumn{1}{l|}{\cellcolor[HTML]{EFEFEF}0,31000} & \multicolumn{1}{l|}{\cellcolor[HTML]{EFEFEF}0,22000} & 0,47000 & \multicolumn{1}{l|}{\cellcolor[HTML]{FFFFFF}0,27465} & \multicolumn{1}{l|}{\cellcolor[HTML]{FFFFFF}0,15568} & \multicolumn{1}{l|}{\cellcolor[HTML]{FFFFFF}0,24440} & \cellcolor[HTML]{CBCEFA}\textbf{0,72771} \\
        \cline{2-10} 
            \multicolumn{1}{|c|}{\multirow{-3}{*}{\cellcolor[HTML]{EFEFEF}\makecell{Mean\\uniformity\\rate (UR)}}} & 2D viewpoint & \multicolumn{1}{l|}{\cellcolor[HTML]{EFEFEF}0,31000} & \multicolumn{1}{l|}{\cellcolor[HTML]{EFEFEF}0,38000} & \multicolumn{1}{l|}{\cellcolor[HTML]{EFEFEF}0,00000} & 0,41000 & \multicolumn{1}{l|}{\cellcolor[HTML]{FFFFFF}0,24951} & \multicolumn{1}{l|}{\cellcolor[HTML]{FFFFFF}0,23182} & \multicolumn{1}{l|}{\cellcolor[HTML]{FFFFFF}0,05559} & \cellcolor[HTML]{CBCEFA}\textbf{0,76610} \\
        \hline
            \multicolumn{2}{c|}{} & \multicolumn{4}{c|}{\cellcolor[HTML]{EFEFEF}Melo et al.} & \multicolumn{4}{c|}{\cellcolor[HTML]{FFFFFF}CP\_HDR} \\
        \cline{3-10} 
    \end{tabular}
    }%
\end{table*}
\begin{table*}[!htb]
    \centering
    \caption{Comparison between Melo et al. and this study using SIFT and SfHDR detectors. The best result for each dataset is highlighted.}
    \label{tab:res-comparison_sift_melo}
    \resizebox{0.8\textwidth}{!}
    {%
    \begin{tabular}{cc|
            >{\columncolor[HTML]{EFEFEF}}l 
            >{\columncolor[HTML]{EFEFEF}}l 
            >{\columncolor[HTML]{EFEFEF}}l 
            >{\columncolor[HTML]{EFEFEF}}l |
            >{\columncolor[HTML]{FFFFFF}}l 
            >{\columncolor[HTML]{FFFFFF}}l 
            >{\columncolor[HTML]{FFFFFF}}l 
            >{\columncolor[HTML]{FFFFFF}}l |}
        \cline{2-10}
            \multicolumn{1}{c|}{} & Detector & \multicolumn{2}{c|}{\cellcolor[HTML]{EFEFEF}\textbf{SIFT}} & \multicolumn{2}{c|}{\cellcolor[HTML]{EFEFEF}\textbf{SfHDR}} & \multicolumn{2}{c|}{\cellcolor[HTML]{FFFFFF}\textbf{SIFT}} & \multicolumn{2}{c|}{\cellcolor[HTML]{FFFFFF}\textbf{SfHDR}} \\
        \cline{2-10} 
            \multicolumn{1}{c|}{} & Database dynamic range & \multicolumn{1}{c|}{\cellcolor[HTML]{EFEFEF}LDR} & \multicolumn{1}{c|}{\cellcolor[HTML]{EFEFEF}HDR} & \multicolumn{1}{c|}{\cellcolor[HTML]{EFEFEF}LDR} & \multicolumn{1}{c|}{\cellcolor[HTML]{EFEFEF}HDR} & \multicolumn{1}{c|}{\cellcolor[HTML]{FFFFFF}LDR} & \multicolumn{1}{c|}{\cellcolor[HTML]{FFFFFF}HDR} & \multicolumn{1}{c|}{\cellcolor[HTML]{FFFFFF}LDR} & \multicolumn{1}{c|}{\cellcolor[HTML]{FFFFFF}HDR} \\
        \hline
            \multicolumn{1}{|c|}{\cellcolor[HTML]{EFEFEF}} & 2D distance & \multicolumn{1}{l|}{\cellcolor[HTML]{EFEFEF}0,00000} & \multicolumn{1}{l|}{\cellcolor[HTML]{EFEFEF}0,00000} & \multicolumn{1}{l|}{\cellcolor[HTML]{EFEFEF}0,00000} & 0,25000 & \multicolumn{1}{l|}{\cellcolor[HTML]{FFFFFF}0,20143} & \multicolumn{1}{l|}{\cellcolor[HTML]{CBCEFA}\textbf{0,26066}} & \multicolumn{1}{l|}{\cellcolor[HTML]{FFFFFF}0,15981} & 0,07161 \\
        \cline{2-10} 
            \multicolumn{1}{|c|}{\cellcolor[HTML]{EFEFEF}} & 2D lighting & \multicolumn{1}{l|}{\cellcolor[HTML]{EFEFEF}0,00000} & \multicolumn{1}{l|}{\cellcolor[HTML]{EFEFEF}0,00000} & \multicolumn{1}{l|}{\cellcolor[HTML]{EFEFEF}0,00000} & \cellcolor[HTML]{CBCEFA}\textbf{0,34000} & \multicolumn{1}{l|}{\cellcolor[HTML]{FFFFFF}0,04101} & \multicolumn{1}{l|}{\cellcolor[HTML]{FFFFFF}0,01314} & \multicolumn{1}{l|}{\cellcolor[HTML]{FFFFFF}0,01552} & 0,00981 \\
        \cline{2-10} 
            \multicolumn{1}{|c|}{\multirow{-3}{*}{\cellcolor[HTML]{EFEFEF}\makecell{Mean\\repeatability\\(sRR)}}} & 2D viewpoint & \multicolumn{1}{l|}{\cellcolor[HTML]{EFEFEF}0,00000} & \multicolumn{1}{l|}{\cellcolor[HTML]{EFEFEF}0,00000} & \multicolumn{1}{l|}{\cellcolor[HTML]{EFEFEF}0,00000} & \cellcolor[HTML]{CBCEFA}\textbf{0,29000} & \multicolumn{1}{l|}{\cellcolor[HTML]{FFFFFF}0,14597} & \multicolumn{1}{l|}{\cellcolor[HTML]{FFFFFF}0,04487} & \multicolumn{1}{l|}{\cellcolor[HTML]{FFFFFF}0,03022} & 0,01519 \\
        \hline
        \hline
            \multicolumn{1}{|c|}{\cellcolor[HTML]{EFEFEF}} & 2D distance & \multicolumn{1}{l|}{\cellcolor[HTML]{EFEFEF}0,28000} & \multicolumn{1}{l|}{\cellcolor[HTML]{EFEFEF}0,43000} & \multicolumn{1}{l|}{\cellcolor[HTML]{EFEFEF}0,02000} & 0,62000 & \multicolumn{1}{l|}{\cellcolor[HTML]{FFFFFF}0,24148} & \multicolumn{1}{l|}{\cellcolor[HTML]{FFFFFF}0,43342} & \multicolumn{1}{l|}{\cellcolor[HTML]{FFFFFF}0,35257} & \cellcolor[HTML]{CBCEFA}\textbf{0,83200} \\
        \cline{2-10} 
            \multicolumn{1}{|c|}{\cellcolor[HTML]{EFEFEF}} & 2D lighting & \multicolumn{1}{l|}{\cellcolor[HTML]{EFEFEF}0,25000} & \multicolumn{1}{l|}{\cellcolor[HTML]{EFEFEF}0,29000} & \multicolumn{1}{l|}{\cellcolor[HTML]{EFEFEF}0,13000} & 0,64000 & \multicolumn{1}{l|}{\cellcolor[HTML]{FFFFFF}0,16073} & \multicolumn{1}{l|}{\cellcolor[HTML]{FFFFFF}0,36771} & \multicolumn{1}{l|}{\cellcolor[HTML]{FFFFFF}0,29542} & \cellcolor[HTML]{CBCEFA}\textbf{0,78000} \\
        \cline{2-10} 
            \multicolumn{1}{|c|}{\multirow{-3}{*}{\cellcolor[HTML]{EFEFEF}\makecell{Mean\\uniformity\\rate (UR)}}} & 2D viewpoint & \multicolumn{1}{l|}{\cellcolor[HTML]{EFEFEF}0,29000} & \multicolumn{1}{l|}{\cellcolor[HTML]{EFEFEF}0,41000} & \multicolumn{1}{l|}{\cellcolor[HTML]{EFEFEF}0,00000} & 0,64000 & \multicolumn{1}{l|}{\cellcolor[HTML]{FFFFFF}0,14800} & \multicolumn{1}{l|}{\cellcolor[HTML]{FFFFFF}0,45657} & \multicolumn{1}{l|}{\cellcolor[HTML]{FFFFFF}0,35285} & \cellcolor[HTML]{CBCEFA}\textbf{0,83285} \\
        \hline
            \multicolumn{2}{c|}{} & \multicolumn{4}{c|}{\cellcolor[HTML]{EFEFEF}Melo et al.} & \multicolumn{4}{c|}{\cellcolor[HTML]{FFFFFF}CP\_HDR} \\
        \cline{3-10} 
    \end{tabular}
    }%
\end{table*}

Comparing the detection results obtained using the CP\_HDR library with Melo et al.~\cite{welerson_hdr}, we observed a considerable improvement in the UR metric. This means that the CP\_HDR library detects more FPs across the image's brightest, intermediate, and darkest areas, while the Melo et al. implementation detected FPs only in the brightest and darkest areas, with no FP detected in the intermediate area.

\section{Conclusions}~\label{sec:conclusion}

In this study, we presented an investigation about the use of HDR images in FP detection and description. Initially, we made a systematic review with 21 studies. Only one study used HDR images, while the others used tone mapping algorithms to convert HDR images into LDR. Two studies presented datasets that explore images with extreme lighting conditions, providing capture versions in LDR and HDR and metrics to evaluate detectors and descriptors.
     
In the previous implementation by Melo et al.~\cite{welerson_hdr}, we have noticed some issues with code maintainability. So, we decided to rewrite it in order to provide a library called CP\_HDR. We implemented the CP\_HDR library to support both LDR and HDR images. It contains the complete SIFT algorithm (with subpixel precision, rotation invariance, and SIFT descriptor), the HfHDR and SfHDR proposed by Melo et al.~\cite{welerson_hdr}. In addition, new detector and descriptor algorithms can easily be added. We also implemented two image descriptors using Harris detector with SIFT descriptor (Harris+SIFT and HfHDR+SIFT), and an algorithm to segment images into groups of lighting intensities, that can be used to partition any image. All methods are documented, and there is a \textit{Makefile} with examples of calls to execute the demonstration programs. The library is available on GitHub~\footnote{Available at: \url{https://github.com/ddantas-ufs/2024_cp_hdr}. (Accessed on July 01, 2023)}. 
    
We observed that CP\_HDR algorithms improved detection results when compared to Melo et al., especially with UR metric. Furthermore, most of the best results were obtained when using HDR images. This is a good indication that the use of HDR images provides better FP detection and description. HfHDR and SfHDR algorithms showed better UR values in most cases.
     
The CP\_HDR library fulfills its purpose, showing improvement in UR and RR when compared to previous studies. HfHDR and SfHDR algorithms presented better detection performance when HDR images are used. The detected FPs are well distributed in all darkest, intermediate, and brightest areas of the image. In the description, SIFT algorithm performed better than the SfHDR. Thus, we consider that SfHDR and HfHDR achieve their goals as they improve FP detection in low-light areas. As future work, we may implement the SURF algorithm in CP\_HDR, which was the one of the most used algorithm in the studies of the systematic review. We may also compare the use of LDR and HDR images with TM-LDR images and improving the CV filter and logarithmic mapping to generate less noise, especially in low-light areas and specular object surfaces.

\section*{Conflict of interest}
    The authors declare that they have no conflict of interest.

\section*{Data availability}
    The code developed during the current study is available on GitHub~\footnote{GitHub repository: \url{https://github.com/ddantas-ufs/2024_cp_hdr}.}.


\bibliographystyle{IEEEtran}
\bibliography{main}

\end{document}